\documentclass[10pt,twocolumn,letterpaper]{article}

\usepackage{iccv}
\usepackage{times}
\usepackage{epsfig}
\usepackage{graphicx}
\usepackage{grffile}
\usepackage{amsmath}
\usepackage{amssymb}
\usepackage{adjustbox}
\usepackage{paralist}

\usepackage[pagebackref=true,breaklinks=true,citecolor=blue,linkcolor=blue,letterpaper=true,colorlinks,bookmarks=false]{hyperref}

\iccvfinalcopy % *** Uncomment this line for the final submission

\def\iccvPaperID{****} % *** Enter the ICCV Paper ID here

\ificcvfinal\fi
\begin{document}

%%%%%%%%% TITLE
\title{Scene Parsing with Global Context Embedding\vspace{-5mm}}

\author{Wei-Chih Hung$^1$, Yi-Hsuan Tsai$^1$, Xiaohui Shen$^2$, \vspace{1mm}\\ 
Zhe Lin$^2$,  Kalyan Sunkavalli$^2$, Xin Lu$^2$, Ming-Hsuan Yang$^1$\vspace{1mm}\\
$^1$University of California, Merced \hspace{20pt} $^2$Adobe Research
}

\maketitle

\vspace{-2mm}

%%%%%%%%% ABSTRACT
\begin{abstract}
We present a scene parsing method that utilizes global context information based on both the parametric and non-parametric models. 
Compared to previous methods that only exploit the local relationship between objects, we train a context network based on scene similarities to generate feature representations for global contexts. 
In addition, these learned features are utilized to generate global and spatial priors for explicit classes inference. 
We then design modules to embed the feature representations and the priors into the segmentation network as additional global context cues.
We show that the proposed method can eliminate false positives that are not compatible with the global context representations.
Experiments on both the MIT ADE20K and PASCAL Context datasets show that the proposed method performs favorably against existing methods.
      
\end{abstract}
\vspace{-4mm}
   
%%%%%%%%% BODY TEXT
\section{Introduction}
\begin{figure}[t]
   \centering\
   \begin{tabular}{cc}
      \multicolumn{2}{c}{\includegraphics[width=0.95\linewidth]{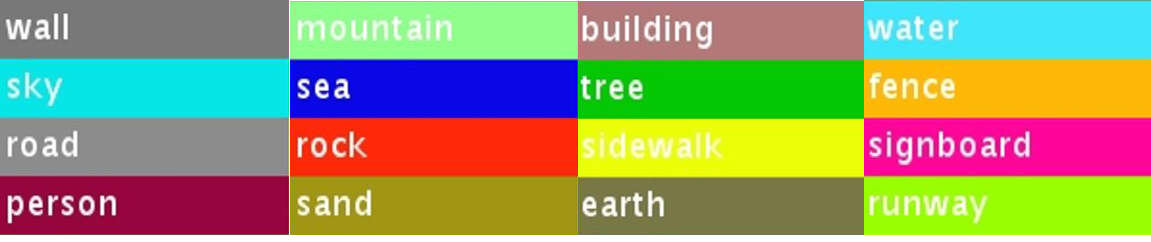}} \\ 
      \includegraphics[width=0.45\linewidth, height=0.3\linewidth]{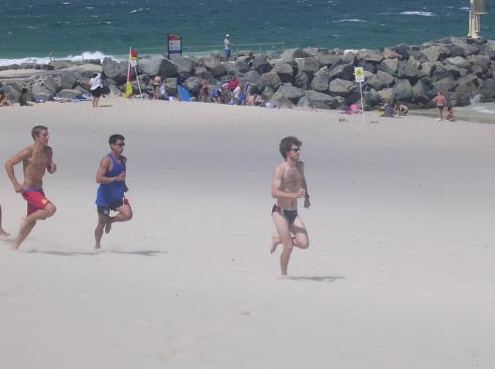} & 
      \includegraphics[width=0.45\linewidth, height=0.3\linewidth]{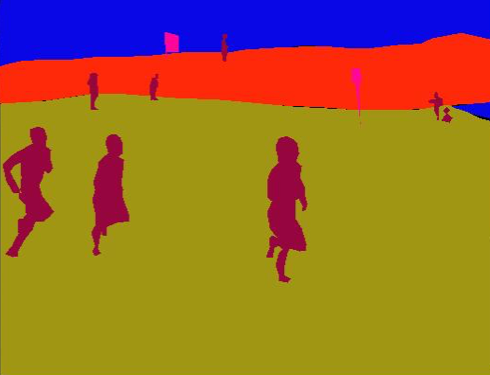} \\
      (a) input & (b) ground truth \\
      \includegraphics[width=0.45\linewidth, height=0.3\linewidth]{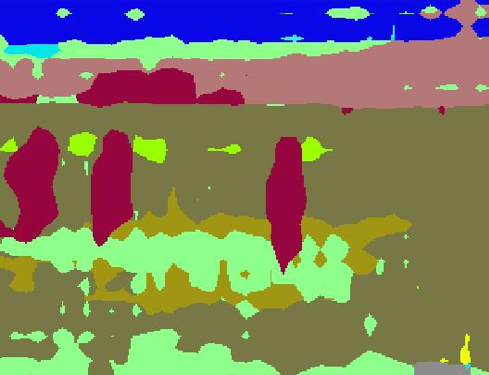} & 
      \includegraphics[width=0.45\linewidth, height=0.3\linewidth]{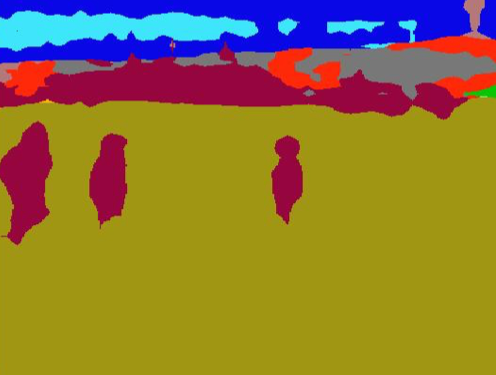} \\
      (c) without global context & (d) with global context \\
      \vspace{0.02cm}
   \end{tabular}
   \caption{
      Given an image (a), the proposed method improves the results of FCN-8s~\cite{fcn_pami} (c)  by exploiting the global context information. Our result (d) shows that the algorithm can eliminate the false positives that are not compatible with the scene category (e.g., some sand regions in (c) are predicted as mountain in a beach scene). (b) shows the ground truth pixel labels.
   }
   \label{figure: idea}
   \vspace{-3mm}
\end{figure}
   
%\paragraph{describe scene parsing}
   
Scene parsing is one of the fundamental and challenging problems in computer vision, 
which can be applied to a wide range of applications 
such as autonomous driving~\cite{cityscapes} and image editing~\cite{tsai2016sky}.
The goal of this task is to assign a semantic class to each pixel in an image. 
Different from semantic segmentation where a significant amount of pixels are labeled as the background, most pixels in the scene parsing datasets are labeled with either 
{\em thing} classes (e.g., person and car) or {\em stuff} classes (e.g., wall, ground, and field).

   %\paragraph{how existing works solve it/their drawback}
   
One major limitation of existing scene parsing methods is that local information only provides limited cues for inferring the label of a single pixel or patch. 
For example, in Figure \ref{figure: idea}(a), when observing a patch filled with gray pixels from the beach sand, it is difficult to infer whether the patch belongs to the class of road, wall, or sand, even for human eyes.
Thus, most existing context-based methods combine one or multiple models that can perform long-range inference through pairwise relationships, e.g., 
Markov Random Field (MRF), Conditional Random Field (CRF), or global attributes
such as scene categories and spatial locations~\cite{context08, co-occurrence}
Non-parametric methods~\cite{siftflow, superparsing, np12, np13, Tighe13,rare} can be seen as context models focusing on image matching via feature descriptors. 
By retrieving a small set of similar images from the annotated dataset, these
methods construct dense correspondences between the input image and the retrieved images on the pixel or superpixel level.
A final prediction map can be obtained through a simple voting or solving an MRF model. 
%
%These algorithms perform well on the benchmark datasets, such as SIFT-Flow~\cite{siftflow} and LM+SUN~\cite{superparsing}. 
%
However, the performance of such exemplar-based approaches highly depends on the quality of the image retrieval module based on hand-crafted features. 
If the retrieved set does not contain one or more semantic classes of the input image, these non-parametric approaches are not expected to parse the scenes well.
% prediction results would be severely degraded.

Recently, CNN based methods such as the fully convolutional neural network (FCN)~\cite{fcn} have achieved the state-of-the-art results in semantic segmentation.
These algorithms improve the scene parsing task compared to conventional non-parametric approaches. 
The performance gain mainly comes from multiple convolutional and non-linear activation layers that can learn data-specific local features to classify each pixel on a local region (i.e., receptive field). 
However, most FCN-based methods still do not utilize an explicit context model, and the receptive field of one pixel classifier is fixed in a given network architecture.

   %\paragraph{what we did here/how our method tackle related works' drawback}

In this paper, we propose an algorithm to embed global contexts into the segmentation network by feature learning and non-parametric prior encoding. 
Different from previous approaches that only consider the contexts within the input image, our method exploits the scene similarities between images without knowing scene categories. 
The key idea is to use a Siamese network~\cite{siamese} for learning global context representations in an unsupervised manner. 
   
We then propose to use the learned representations to exploit global contexts using global context feature encoding and non-parametric prior encoding. 
For global context feature encoding, we propagate the learned representations through the segmentation network. 
For non-parametric prior encoding, we generate both global and spatial priors by retrieving annotated images via global context features and combine them with our segmentation network.
%
%Our non-parametric model is efficient regarding both computation time and memory usage. It is because we do not perform dense alignment like most existing non-parametric methods, and the retrieval process can be efficiently done by comparing the global feature with stored compact feature representations of the training dataset.
Since we do not perform dense alignment as most existing non-parametric methods~\cite{siftflow, superparsing}, our non-parametric module is computationally efficient. Instead of using the original images from the whole training set, which requires large storage at the testing phase, our image retrieval module only needs the pre-computed compact feature representations of images.
We evaluate the proposed algorithm on the MIT ADE20K~\cite{zhou2016semantic} and PASCAL Context dataset~\cite{pascal_context} with comparisons to the state-of-the-art methods.
%
%The results show that the proposed method achieves favorable performance.
   
The contributions of this work are as follows:
%\begin{enumerate}
\begin{compactitem}
\item We design a Siamese network to learn representations for global contexts and model scene similarities between images in an unsupervised manner, 
with a focus on images that share rare object/surface categories.
      
\item We propose two methods to exploit global contexts by feature learning and non-parametric prior encoding.
      
\item We show that the parametric segmentation network, context encoding, and non-parametric prior encoding can be efficiently integrated via 
the proposed global context embedding scheme
for effective scene parsing without introducing much computational overhead.
      
%\end{enumerate}
\end{compactitem}

\section{Related Work}
   
   %\paragraph{methods with context information}
   
Numerous methods have been developed to exploit scene contexts for vision tasks.
In~\cite{context08}, the Thing and Stuff model that exploits the relationship between objects and surfaces is proposed to eliminate false positives of object detectors. 
%
%A similar idea is proposed in the form of co-occurrence~\cite{co-occurrence} and addressed by the MRF optimization.
%
In~\cite{rare, np12}, the initial prediction of adjacent superpixels is used as the local context descriptor to refine the superpixel matching process iteratively.
Most methods based on FCN~\cite{fcn} exploit context information by constructing MRFs or CRFs on top of the network output~\cite{crfasrnn, deeplab, deepparsing, piecewise, segDeepM}.
A recent study shows that the performance of FCN-based models can be improved by applying dilated convolution~\cite{dilated}.
The key insight is that dilated convolution allows the network to ``see'' more, i.e., enlarging the receptive field, and therefore more context information is perceived.
However, these models only consider the local context information within the input image. 
Our proposed context model is related to the global context descriptors in~\cite{rare, np13} which refine the scene retrieval module in the non-parametric label transfer. 
In this work, we use the global descriptor to improve the pixel classifier directly. 
Recently, one concurrent work~\cite{zhao2016pyramid} proposes to exploit the global context through pyramid pooling. Different from \cite{zhao2016pyramid}, we train our context network with an explicit distance metric.
   
\begin{figure*}[ht]
\centering
\includegraphics[width=\linewidth]{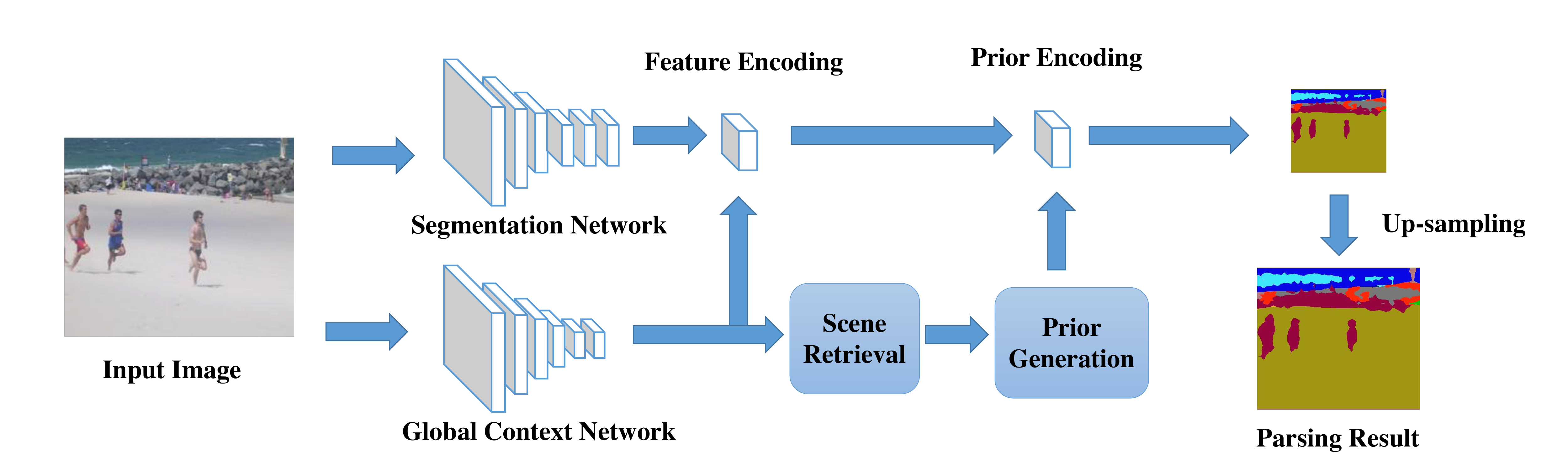}\\
\caption{Algorithmic overview. 
We propagate an input image through two networks: segmentation network for initial local label prediction and global context network for generating global context features. 
We exploit the learned features with context feature encoding and non-parametric prior encoding. 
These two modules can be easily applied to existing segmentation networks. 
After applying our methods to the segmentation network, the final scene parsing results are obtained through a softmax operation and a bilinear upsampling.}
      \label{figure: system}
\vspace{-3mm}
\end{figure*}
   
   %\paragraph{non-parametric works}
   % global feature, alignment

Our prior encoding approach is closely related to the non-parametric approaches for scene parsing ~\cite{siftflow, superparsing, np12, np13, Tighe13, rare}. 
These methods typically consist of three major stages: scene retrieval, dense alignment, and MRF/CRF optimization. 
In the scene retrieval stage, a global descriptor is used to retrieve a small set of images from the annotated dataset based on different features, e.g., 
GIST and HoG features~\cite{siftflow}, dense-SIFT~\cite{rare}, superpixels  and hybrid features~\cite{superparsing, np12, np13, Tighe13}.
%  First stage prediction result is also used to improve retrieval stage in \cite{np13, rare}.
However, these hand-crafted features are less effective in describing small objects. 
In this work, we propose to replace the global descriptor with the deep features trained by the Siamese network~\cite{siamese} to enhance the semantic embedding. 
The dense alignment can be achieved via the SIFT-Flow~\cite{siftflow}, superpixel matching~\cite{superparsing, rare, np12, np13}, or exemplar-based classifier~\cite{Tighe13}. 
However, the explicit dense alignment requires heavy computational loads.
%
%In this work, we do not perform dense alignment explicitly since it requires heavy computational loads. 
%
In this work, we generate the prior information without alignment and pass the priors through convolutional layers. 
It allows the segmentation network to learn how to combine the prior information with local prediction in an end-to-end fashion. 
%

   %\paragraph{siamese network}
   
As a core module in our method, the Siamese network~\cite{siamese} can learn the pair-wise relationship between images. 
The network transforms a classification network to multiple branches with shared parameters. 
Such network structures have been used for re-identification problems ~\cite{yi2014deep, ahmed2015improved} and unsupervised visual learning~\cite{deepcontext, wang2015unsupervised, featurelearning}.

\section{Algorithmic Overview}
   
% describe the system flow
   
Figure \ref{figure: system} shows the overview of the proposed algorithm. 
We first propagate an input image through two networks: segmentation network and global context network. 
The segmentation network  (e.g., FCN~\cite{fcn} or DeepLab~\cite{deeplab})
generates the initial parsing results. 
The global context network is designed based on a CNN classification network (e.g., AlexNet~\cite{alexnet} or VGG~\cite{vgg}) without the last fully-connected and softmax layers.
The global context network outputs a fixed length feature vector that embeds the global context information of the input image.
The context embedded features are then combined with the feature maps of the segmentation network, passing through the feature encoding network to exploit additional information. 
   
In addition to feature encoding, we propose a prior encoding module to combine the non-parametric prior information.
We obtain the spatial and global prior information by retrieving the $K$-nearest annotated images from the training set using the learned context embedded features.
These priors estimate the label distribution of the retrieved images.
We then encode the prior information to the segmentation network using the proposed prior encoding network.
To match the size of the input image, we apply the bilinear upsampling on the output of the prior encoding module as the final parsing results.

\section{Training Global Context Network}
\label{section: GCN}
   
To obtain the global context information, a straightforward approach is to generate scene labels on the image level, e.g., bedroom, school, or office. 
However, it requires additional annotations. 
Moreover, unlike object categories, scene categories are often ambiguous, and the boundaries between some categories are difficult to draw, e.g., a scene consists of both the street view and the outdoor dining area.
Thus, instead of explicitly inferring the scene categories, we propose to embed the global context into a fixed-dimensional semantic space through the global context network. 
The objective of the global context network is to capture the scene information of the desired semantic embedding properties. 
For example, the feature distance from a bedroom scene to a living room scene should be smaller than that to a beach scene. 
We observe that semantically similar scenes share more common classes, e.g. wall, chair, and sofa for indoor scenes. 
Toward this end, we design a distance metric, denoted as ground truth distance, to evaluate the semantic distance between a pair of images based on their annotated pixel labels.
The ground truth distance provides rich context information for the scene parsing task, and our objective is to utilize such context information in the testing phase without knowing pixel labels.
Therefore, we propose to train a global context network to generate the global context features by learning from the ground truth distance. We demonstrate that the distances between trained global context features have the similar semantic embedding of the ground truth distance.

% chi-square, spatial pyramid
\vspace{1mm}
{\noindent {\bf Ground Truth Distance.}} The ground truth distance describes the semantic distance between two images with annotated pixel labels.
We denote it as $d_{gt}(\mathbf{y_i}, \mathbf{y_j})$, where $\mathbf{y_i}$ and $\mathbf{y_j}$ are the annotated ground truth labels of two images. 
To compute the ground truth distance, we first construct a spatial pyramid on each annotated image. 
In the spatial pyramid, we compute the Chi-square distance between the label histograms of two corresponding blocks at the same location from two images. 
We obtain the ground truth distance by summing up the distance of all blocks, i.e.,
\begin{equation}
d_{gt}(\mathbf{y_i}, \mathbf{y_j}) = \sum_{s \in S}\sum_{c \in C} \chi^2(h_{i}(s,c), h_j(s,c)),
\end{equation}
where $h_i(s,c)$ is the number of pixels belonging to class $c$ at location $s$ in the spatial pyramid.
   
The purpose of constructing the spatial pyramid is to estimate the scene similarity between images with consideration of the spatial scene layout. 
In this work, we use a two-level spatial pyramid where the first level contains only one block and the second level contains $9$ blocks by dividing the image into a $3\times3$ grid. 
We observe that there is no significant difference with more levels of the spatial pyramid. 
We choose the Chi-square distance defined by $\chi^2(a,b) = (a-b)^2/(a+b)$ for computing the distance between histograms since the normalization term can remit the situation that major classes with a large amount of pixels dominate the distance function.
   
\begin{figure}[t]
   \centering
   \includegraphics[width=\linewidth]{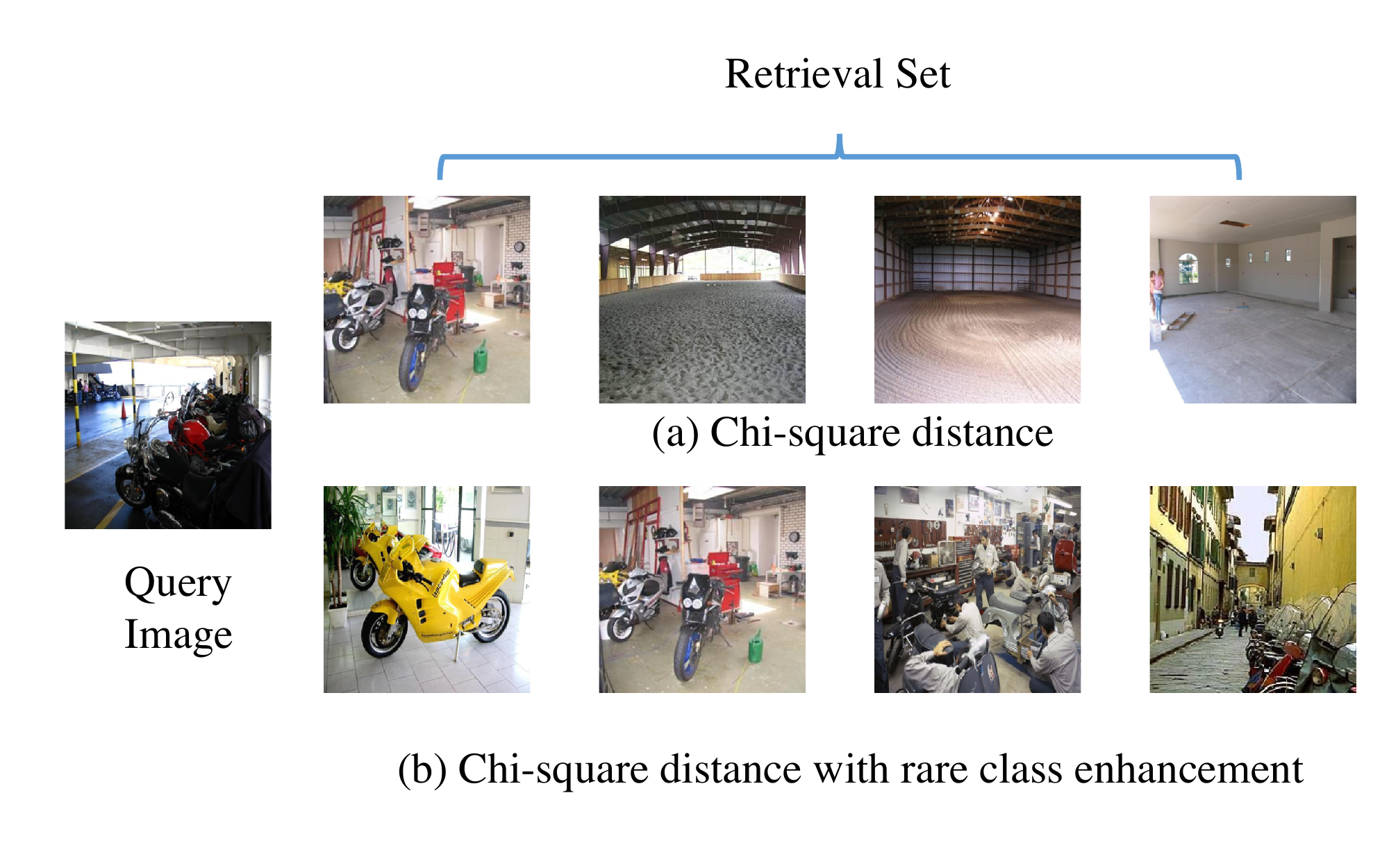}
   \vspace{-4mm}
   \caption{We refine the ground truth distance metric through rare class enhancement. Given a query image, we show the retrieval set obtained by using the ground truth distance (a), and with rare class enhancement (b). 
   Samples of rare classes can be better retrieved with the enhanced metric. 
   %In this example, the motorbike should be more informative than ground or wall.
   }
   \vspace{-3mm}
   \label{figure: rare_enhance}
   \end{figure}
\vspace{1mm}

{\noindent {\bf Rare Class Enhancement.}}
In large-scale vision tasks, 
e.g., object detection~\cite{ouyang2016factors} and scene parsing~\cite{rare, zhou2016semantic}, 
the distribution of annotated class samples is usually highly unbalanced with a long-tail distribution.
%
%This is observed in many fields, including  
%
The unbalanced samples between classes make most learning based methods 
prone to disregarding the rare/small classes to achieve higher overall accuracy. 
In addition, samples of rare classes only appear in certain specific scene categories, e.g., tent and microwave, and provide strong cues for the global context inference. 
In other words, when samples from a rare class appear in a scene, 
local informative regions should be weighted more than the overall global configuration.
   
We re-weight the histogram $h_i(s,c)$ in the ground truth distance $d_{gt}$ by dividing how often the class $c$ appears in the dataset, i.e.,
\begin{equation}
   h_{i}^{r}(s,c) = h_{i}(s,c) / f(c),
\end{equation}
where $f(c)$ is the amount of images in which class $c$ presents with at least one pixel within the dataset.
Figure \ref{figure: rare_enhance} shows an example of the proposed rare class enhancement by comparing the retrieval results using the ground truth distance.

\begin{figure}[t]
      \centering
      \includegraphics[width=\linewidth]{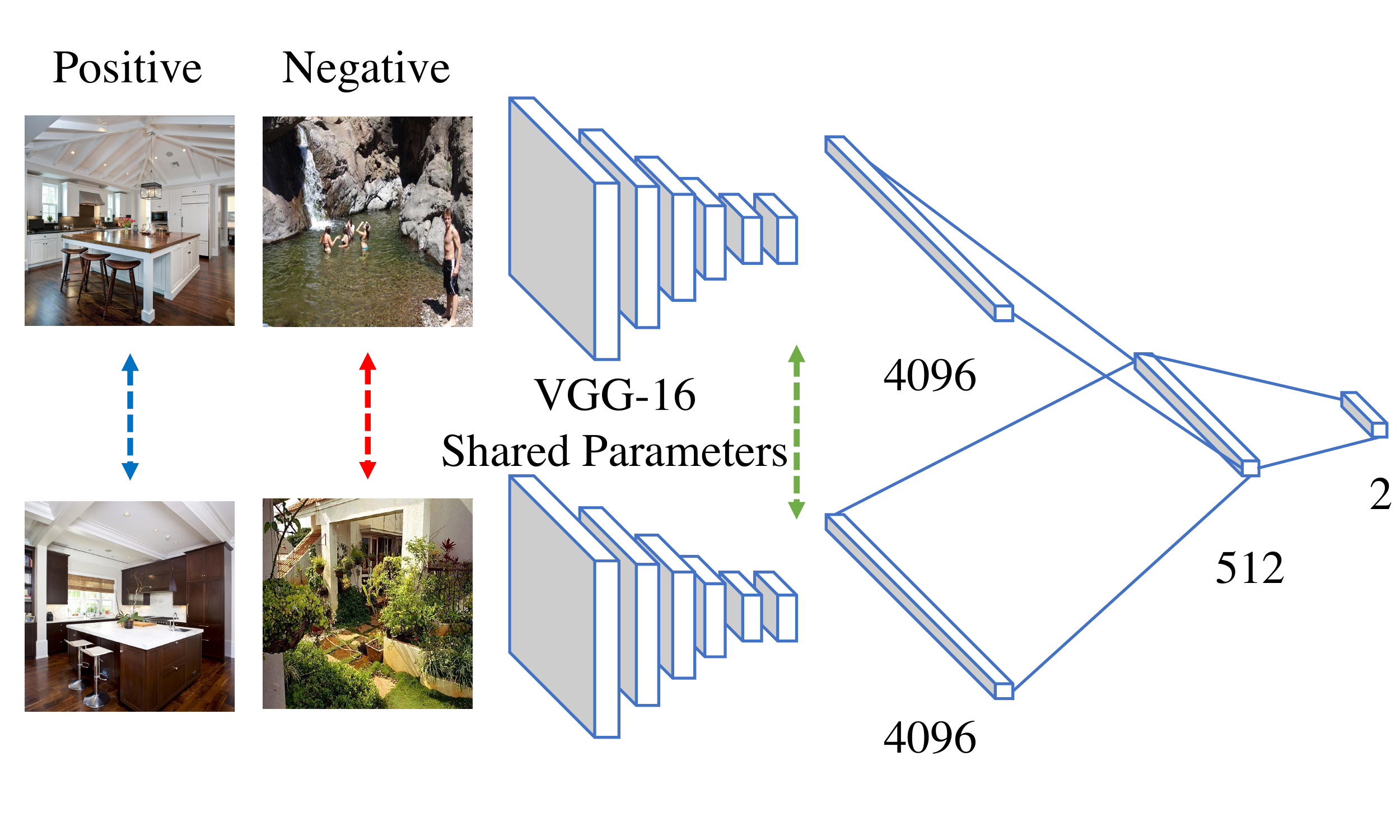}\\
      \caption{Siamese training of the global context network. The global context network is a VGG-16 network that outputs a 
4096-dimensional feature vector. We train the global context network using Siamese structure, in which there are two identical networks with shared parameters. The output feature vectors of the two branches are concatenated and passed through additional fully-connected layers and a softmax layer. The target labels are 1 for positive pairs and 0 for negative pairs.}
      \vspace{-3mm}
      \label{figure: siamese}
\end{figure}

%positive and negative pairs mining
\vspace{1mm}
{\noindent {\bf Siamese Network Training.}}
Given the ground truth distance function $d_{gt}$ between images, we learn a feature space to predict the similar distance embedding. 
Motivated by recent methods that utilize the Siamese network to learn pairwise relationships~\cite{deepcontext, featurelearning, wang2015unsupervised}, 
we train the global context network with a Siamese structure to predict the scene similarity between image pairs, as shown in 
Figure \ref{figure: siamese}.
The design of the global context network is based on the VGG-16~\cite{vgg} model that takes a single image as input.
We remove the last fully-connected layer, making \textbf{fc7} layer as the output feature.
The Siamese network consists of two identical global context branches.
On top of the two branch network, two additional fully-connected layers with softmax output are implemented as a binary classifier.
   
By using Siamese structure training, we transform the distance regression task into a binary classification problem. 
For each pair of input images, the network classifies it as either positive (semantically similar) or negative (semantically distant) pair. 
To extract the positive and negative pairs, we first form a fully-connected graph where each node is an image in the annotated dataset, and the edge weight is the ground truth distance function $d_{gt}(\mathbf{y_i}, \mathbf{y_j})$.

We construct an affinity matrix as
   \begin{equation}
   A_{gt}[i,j] =
   \begin{cases}
   1, & j \in \textit{KNN}(i, K_a) \\
   0, & otherwise
   \end{cases}
   ,
   \label{eqn: binarize}
   \end{equation}
where $\textit{KNN}(i, K_a)$ denotes the set that contains $K_a$-nearest neighbors of node $i$ 
with respect to the the ground truth distance $d_{gt}(\mathbf{y_i}, \mathbf{y_j})$.

\begin{figure}[t]
   \scriptsize
   \centering
   \begin{tabular}{cc}
      \begin{adjustbox}{valign=c}
         \begin{tabular}{c}
            \includegraphics[width=0.25\linewidth, height=0.25\linewidth]{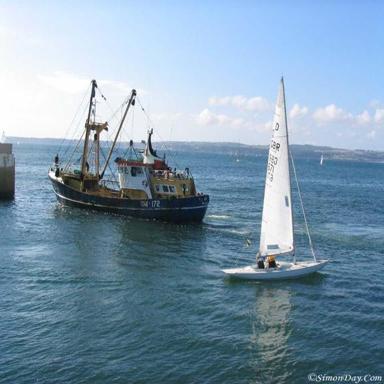}
            \\
            query image
         \end{tabular}
      \end{adjustbox}
      \hspace{-0.32cm}
      \begin{adjustbox}{valign=c}
         \begin{tabular}{cccc}
            \includegraphics[width=0.15\linewidth, height=0.15\linewidth]{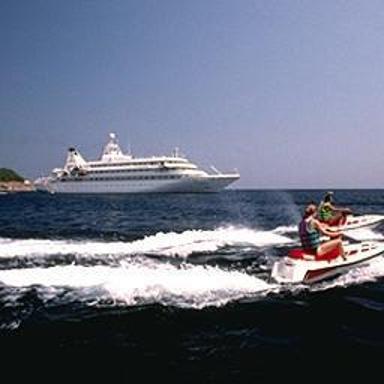} \hspace{-0.3cm} &
            \includegraphics[width=0.15\linewidth, height=0.15\linewidth]{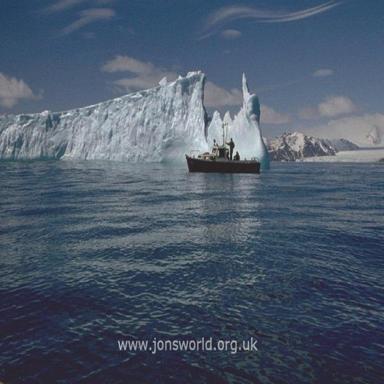} \hspace{-0.3cm} &
            \includegraphics[width=0.15\linewidth, height=0.15\linewidth]{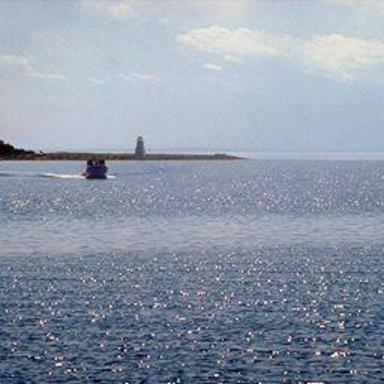} \hspace{-0.3cm} &
            \includegraphics[width=0.15\linewidth, height=0.15\linewidth]{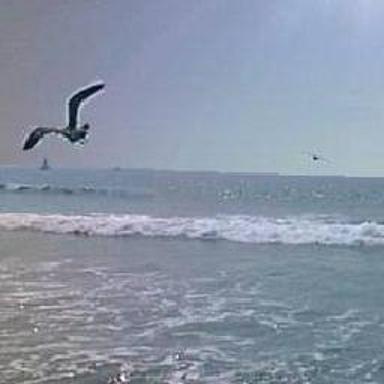}

            \\
            \includegraphics[width=0.15\linewidth, height=0.15\linewidth]{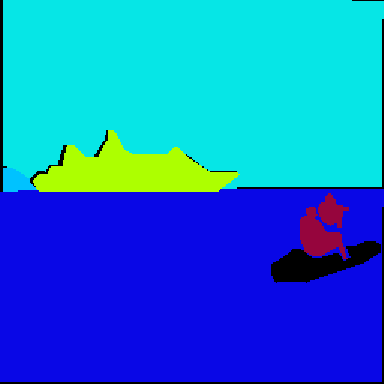} \hspace{-0.3cm} &
            \includegraphics[width=0.15\linewidth, height=0.15\linewidth]{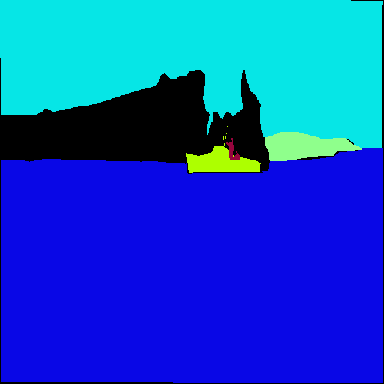} \hspace{-0.3cm} &
            \includegraphics[width=0.15\linewidth, height=0.15\linewidth]{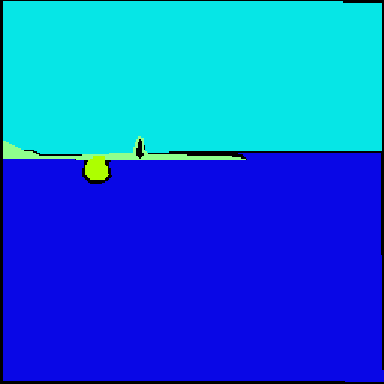} \hspace{-0.3cm} &
            \includegraphics[width=0.15\linewidth, height=0.15\linewidth]{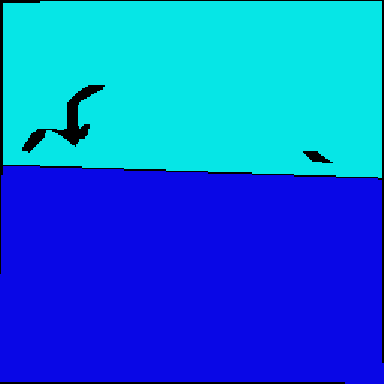} 
            \\ 
            \multicolumn{4}{c}{retrieval set using ground truth distance}
         \end{tabular}
      \end{adjustbox}
   \end{tabular}

   \begin{tabular}{cc}
      \begin{adjustbox}{valign=c}
         \begin{tabular}{c}
            \includegraphics[width=0.25\linewidth, height=0.25\linewidth]{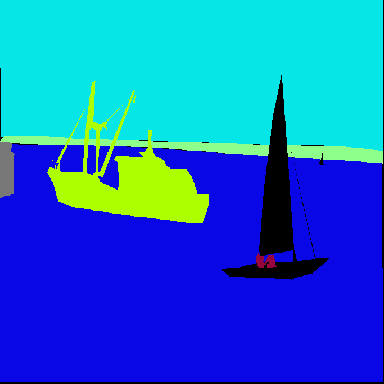}
            \\
            annotation
         \end{tabular}
      \end{adjustbox}
      \hspace{-0.32cm}
      \begin{adjustbox}{valign=c}
         \begin{tabular}{cccc}
            \includegraphics[width=0.15\linewidth, height=0.15\linewidth]{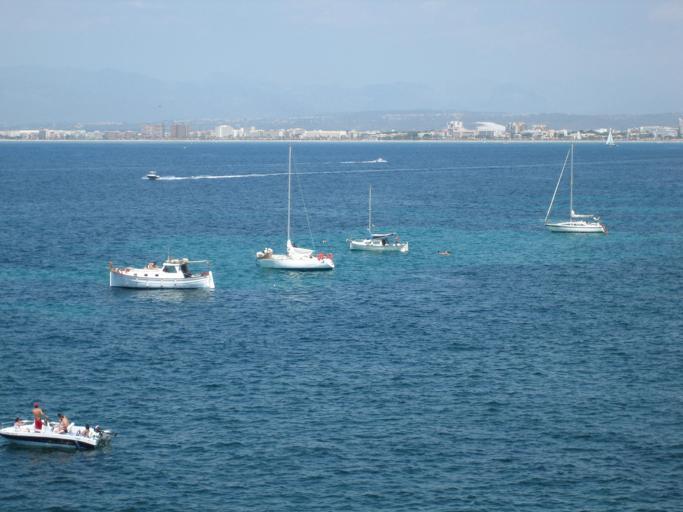} \hspace{-0.3cm} &
            \includegraphics[width=0.15\linewidth, height=0.15\linewidth]{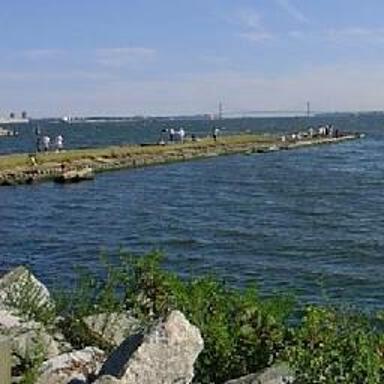} \hspace{-0.3cm} &
            \includegraphics[width=0.15\linewidth, height=0.15\linewidth]{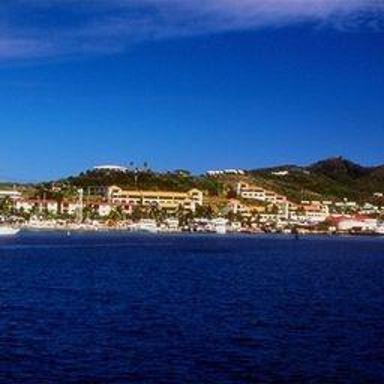} \hspace{-0.3cm} &
            \includegraphics[width=0.15\linewidth, height=0.15\linewidth]{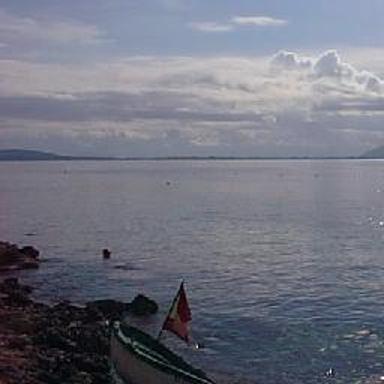}

            \\
            \includegraphics[width=0.15\linewidth, height=0.15\linewidth]{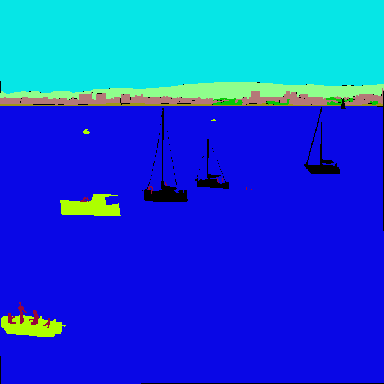} \hspace{-0.3cm} &
            \includegraphics[width=0.15\linewidth, height=0.15\linewidth]{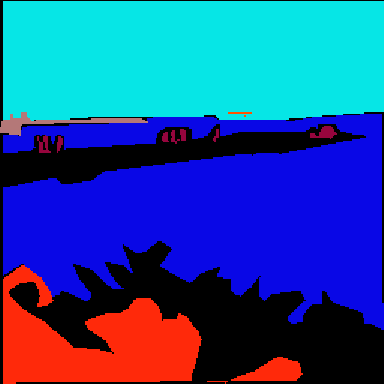} \hspace{-0.3cm} &
            \includegraphics[width=0.15\linewidth, height=0.15\linewidth]{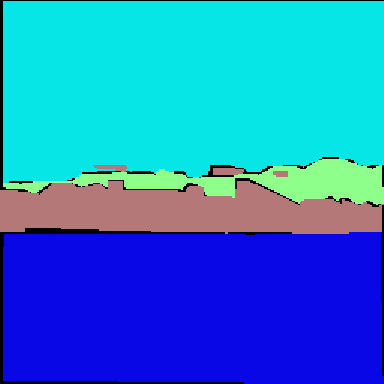} \hspace{-0.3cm} &
            \includegraphics[width=0.15\linewidth, height=0.15\linewidth]{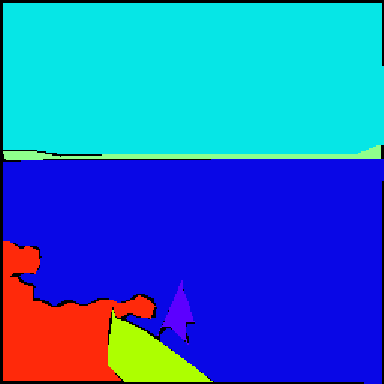} 
            \\ 
            \multicolumn{4}{c}{retrieval set using trained feature distance}
         \end{tabular}
      \end{adjustbox}
   \end{tabular}
   
   \vspace{0.1cm}
   
   \begin{tabular}{cc}
      \begin{adjustbox}{valign=c}
         \begin{tabular}{c}
            \includegraphics[width=0.25\linewidth, height=0.25\linewidth]{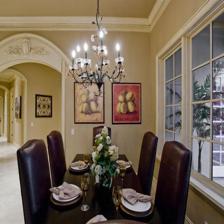}
            \\
            query image
         \end{tabular}
      \end{adjustbox}
      \hspace{-0.32cm}
      \begin{adjustbox}{valign=c}
         \begin{tabular}{cccc}
            \includegraphics[width=0.15\linewidth, height=0.15\linewidth]{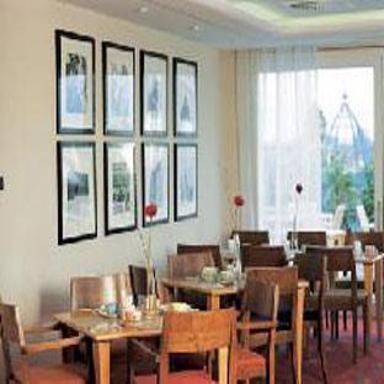} \hspace{-0.3cm} &
            \includegraphics[width=0.15\linewidth, height=0.15\linewidth]{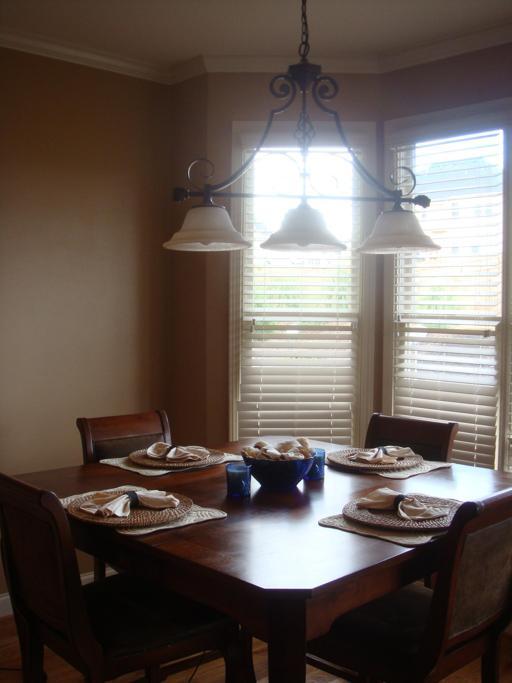} \hspace{-0.3cm} &
            \includegraphics[width=0.15\linewidth, height=0.15\linewidth]{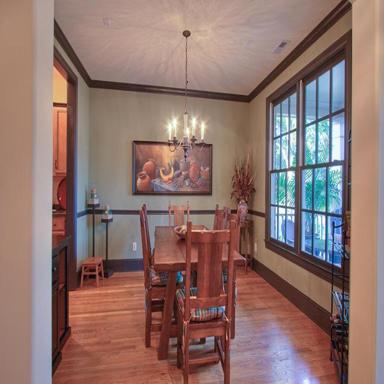} \hspace{-0.3cm} &
            \includegraphics[width=0.15\linewidth, height=0.15\linewidth]{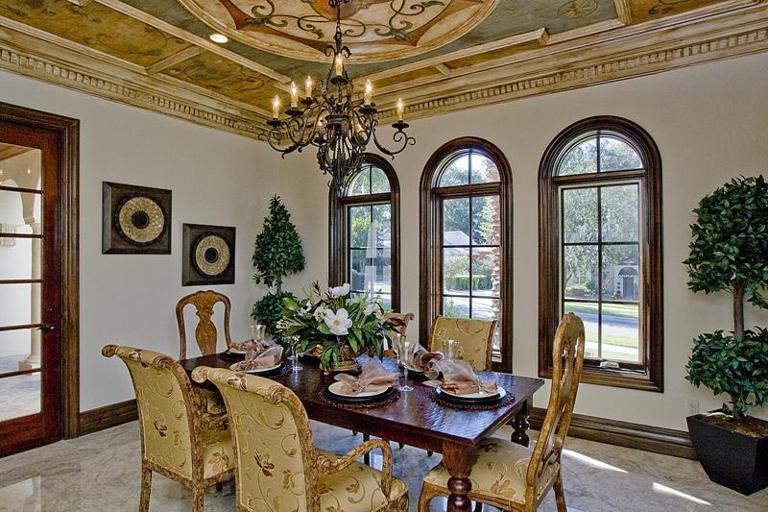}

            \\
            \includegraphics[width=0.15\linewidth, height=0.15\linewidth]{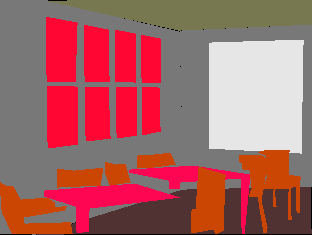} \hspace{-0.3cm} &
            \includegraphics[width=0.15\linewidth, height=0.15\linewidth]{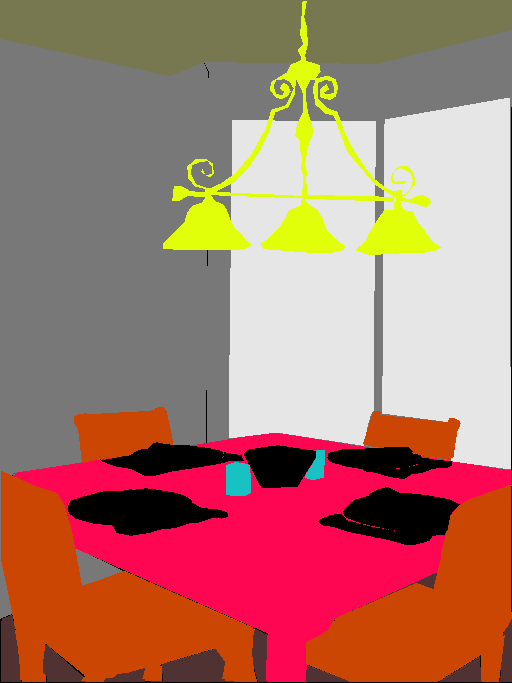} \hspace{-0.3cm} &
            \includegraphics[width=0.15\linewidth, height=0.15\linewidth]{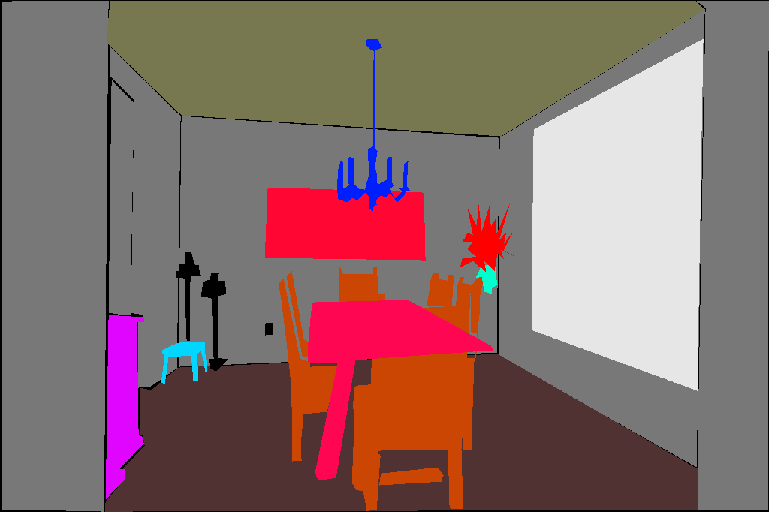} \hspace{-0.3cm} &
            \includegraphics[width=0.15\linewidth, height=0.15\linewidth]{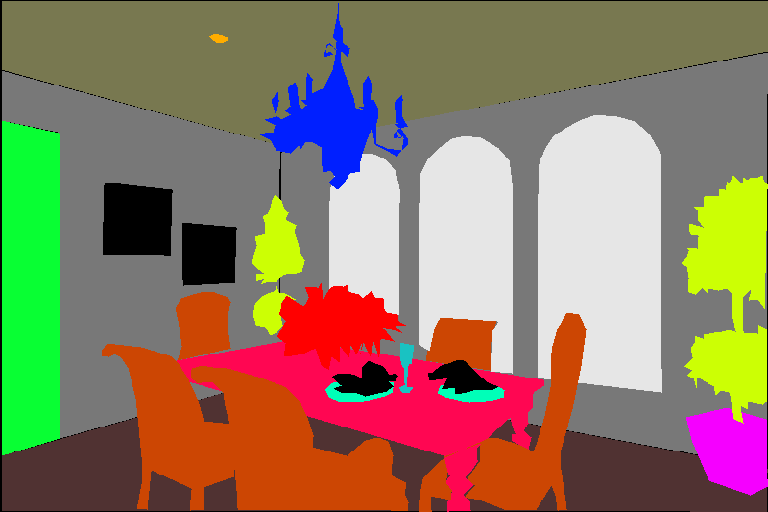} 
            \\ 
            \multicolumn{4}{c}{retrieval set using ground truth distance}
         \end{tabular}
      \end{adjustbox}
   \end{tabular}

   \begin{tabular}{cc}
      \begin{adjustbox}{valign=c}
         \begin{tabular}{c}
            \includegraphics[width=0.25\linewidth, height=0.25\linewidth]{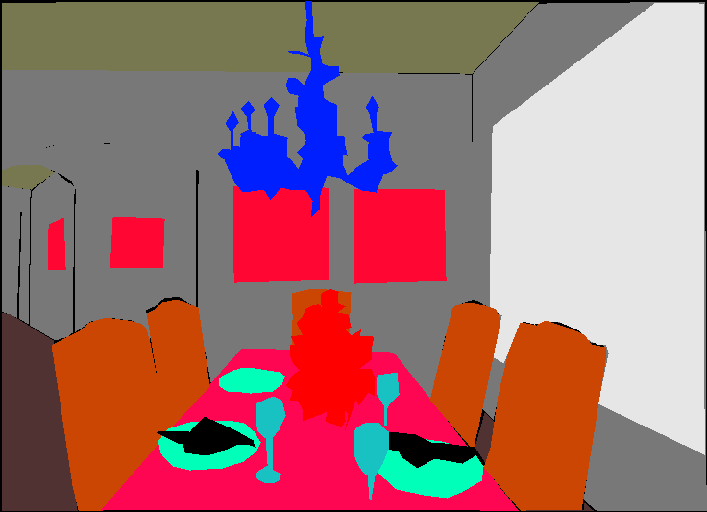}
            \\
            annotation
         \end{tabular}
      \end{adjustbox}
      \hspace{-0.32cm}
      \begin{adjustbox}{valign=c}
         \begin{tabular}{cccc}
            \includegraphics[width=0.15\linewidth, height=0.15\linewidth]{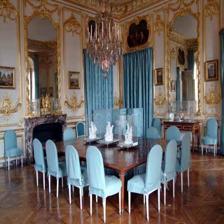} \hspace{-0.3cm} &
            \includegraphics[width=0.15\linewidth, height=0.15\linewidth]{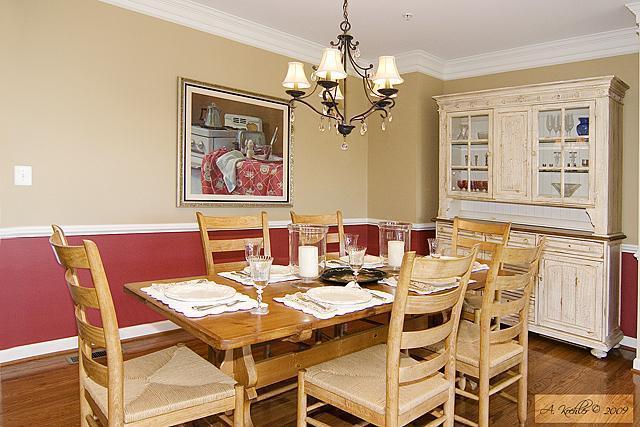} \hspace{-0.3cm} &
            \includegraphics[width=0.15\linewidth, height=0.15\linewidth]{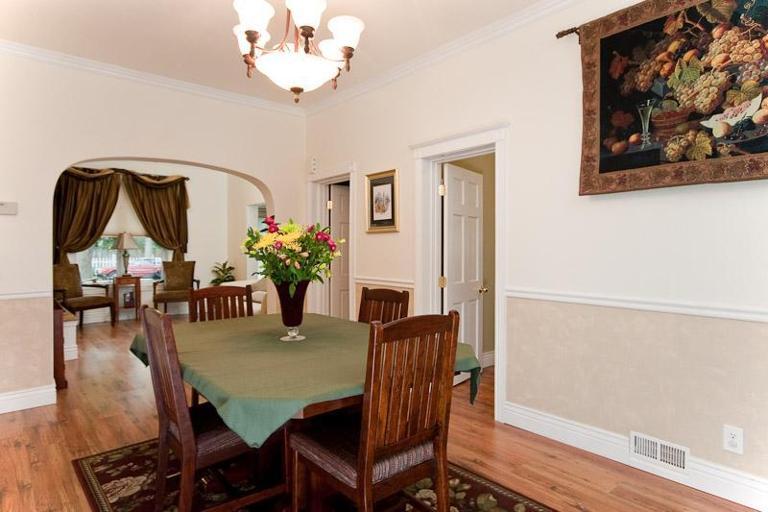} \hspace{-0.3cm} &
            \includegraphics[width=0.15\linewidth, height=0.15\linewidth]{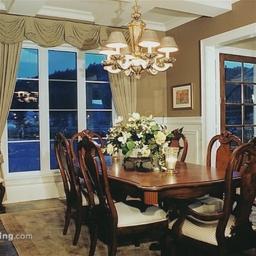}

            \\
            \includegraphics[width=0.15\linewidth, height=0.15\linewidth]{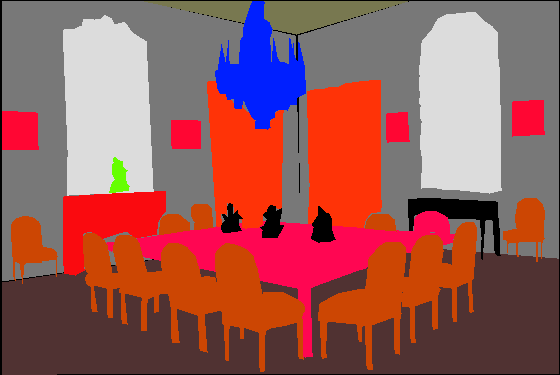} \hspace{-0.3cm} &
            \includegraphics[width=0.15\linewidth, height=0.15\linewidth]{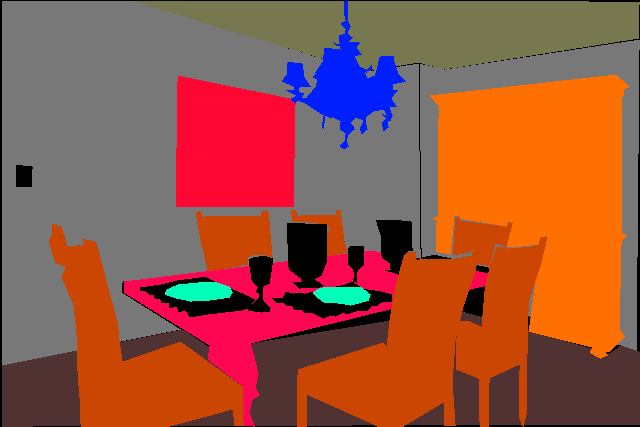} \hspace{-0.3cm} &
            \includegraphics[width=0.15\linewidth, height=0.15\linewidth]{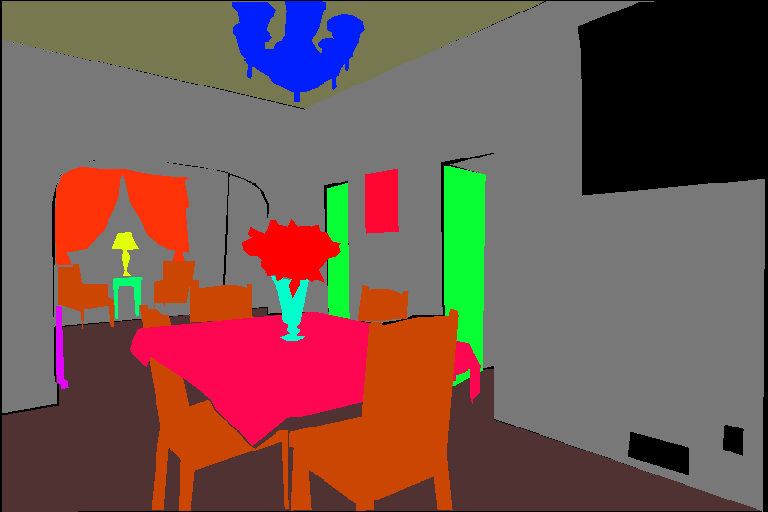} \hspace{-0.3cm} &
            \includegraphics[width=0.15\linewidth, height=0.15\linewidth]{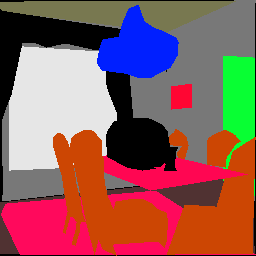} 
            \\ 
            \multicolumn{4}{c}{retrieval set using trained feature distance}
         \end{tabular}
      \end{adjustbox}
   \end{tabular}

   \vspace{0.1cm}
   \caption{ Comparison of retrieval results using ground truth distance and Euclidean distance between trained global context features. The retrieval results using our global context features is simliar to the ones using the designed ground truth distance. }
   \vspace{-4mm}
   \label{figure: siamese_retrieval}
\end{figure}

Since $K_a$ in the nearest neighbor search is relatively small compared to the number of nodes, most entries in $A_{gt}$ are zeros and treated as negative pairs. 
However, it is impractical to train the Siamese network using all the negative pairs. 
Therefore, we need to sample negative pairs during training. 
A straightforward method is random sampling.
Nevertheless, not all the negative pairs are equally informative.
A pair with large ground truth distance can be considered as an easy sample that does not help the network to learn discriminative features.
Toward this end, we apply a simple strategy to mine the hard-negative pairs by only sampling negative pairs from the $K_a+1$ nearest neighbor to $N$-th nearest neighbor, where $N$ is larger than $K_a$. 
In this work, we set the $N$ as half the amount of images in the training dataset.
   
Figure \ref{figure: siamese_retrieval} shows the retrieval results using the ground truth distance and Euclidean distance of trained features. 
The results show that the trained global context features can represent similar semantics from the ground truth distance using the Siamese network training.
   
%==================================================================  
\section{Non-parametric Prior Generation}
\label{section: prior}
%\vspace{1mm}  
{\noindent {\bf Scene Retrieval.}}
Motivated by the non-parametric work on scene parsing~\cite{siftflow, superparsing, np12, np13, Tighe13, rare}, we propose a method to utilize non-parametric prior in the segmentation network.
Given a query image, we use a global descriptor to retrieve a small set of semantically similar images from the training dataset.
Compared to other methods that use hand-crafted features, we use the learned global context features as the image descriptor.
Specifically, given a query image $\mathbf{x}_i$ with its global context features $\mathbf{f}_i$, we retrieve an image set $\{\mathbf{x}_1, \mathbf{x}_2, \ldots, \mathbf{x}_{K_p}\}$ by performing ${K_p}$-nearest-neighbors search 
with the Euclidean distance between the context features.
Note that we use $K_p$ here to differentiate with the $K_a$ in the Siamese training.
Figure \ref{figure: siamese_retrieval} also shows some example retrieval sets using both the ground truth distance and the Euclidean distance between trained features.
Since the retrieval images are semantically close to the query image, the annotated pixel labels in the retrieval set should also be similar to the query image and thus can be used to help the parsing process of the query image.
   
While most previous methods use dense alignment to transfer the retrieved labels as the parsing result directly, we propagate the prior probability through a convolution layer to jointly predict the results with the segmentation network.
To design the prior, we observe that the {\em stuff} classes such as sky and ground do have strong spatial dependency (i.e., sky is usually on the top and ground at the bottom of most images), 
while {\em things} classes such as chair and person can appear at most image locations depending on the camera view angle. 
Therefore, we propose to exploit two forms of prior information: spatial prior and global prior.
The spatial prior estimates how likely the {\em stuff} classes presenting at each spatial location, while global prior estimates the probability of existence for {\em things} classes on the image level.

\begin{figure*}[t]
   \footnotesize
   \footnotesize
   \centering
   \begin{tabular}{cc}
      \includegraphics[width=0.45\linewidth]{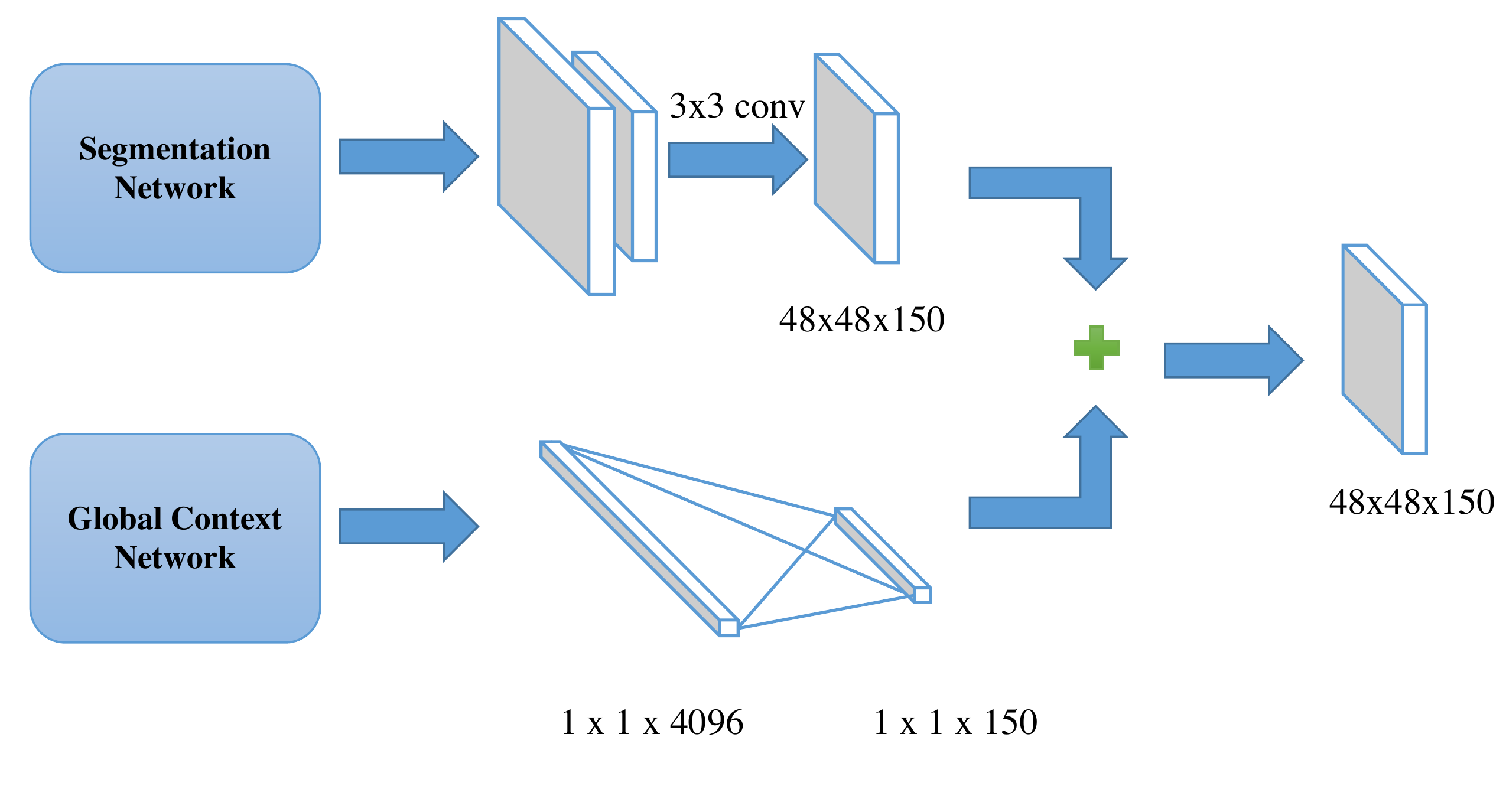} &
      \includegraphics[width=0.45\linewidth]{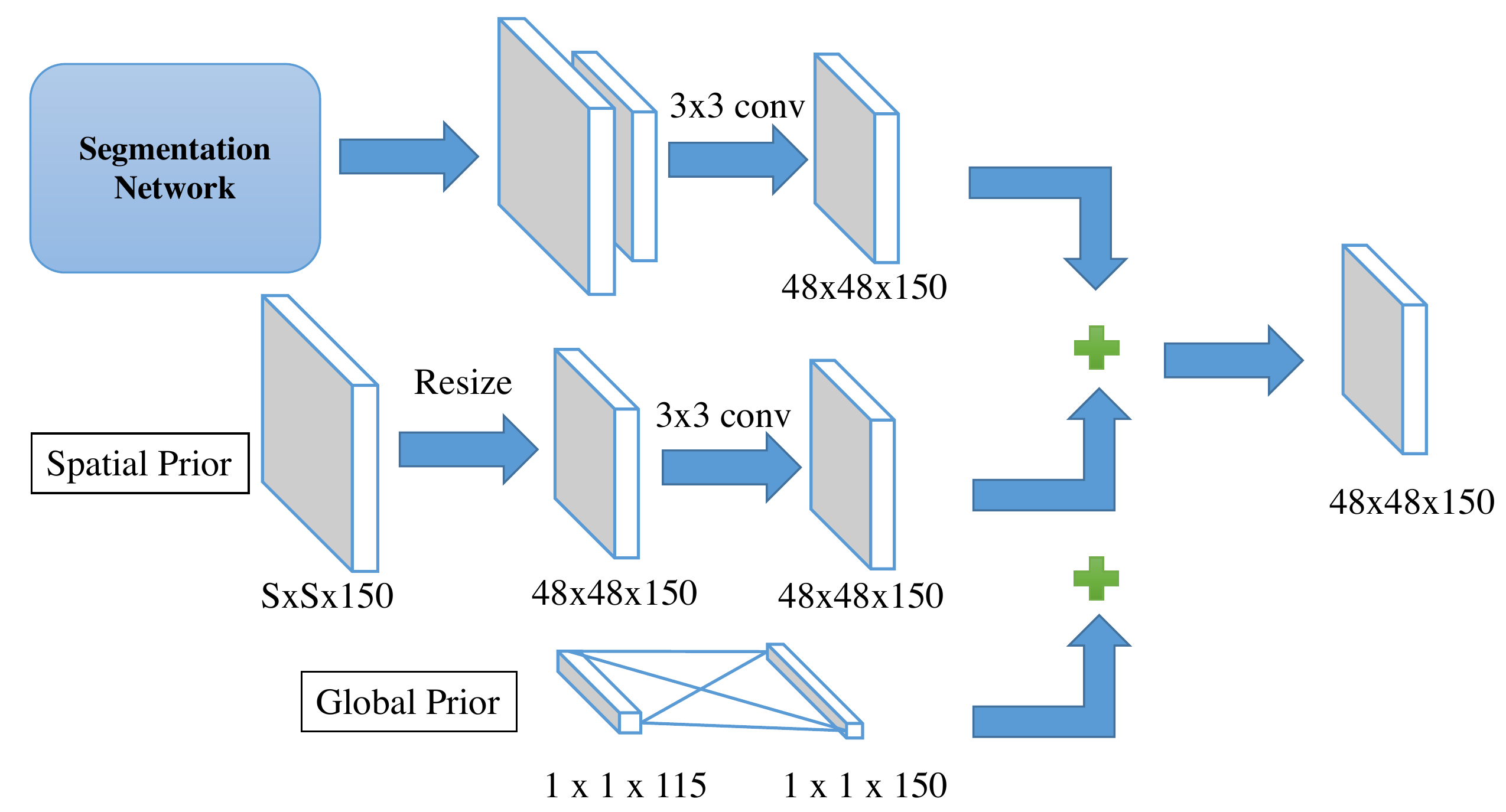}\\
      (a) Feature Encoding &
      (b) Prior Encoding\\
   \end{tabular}
   \vspace{0.1cm}
   \caption{Proposed encoding structures. (a) shows the structure to encode the features generated by the global context network. We first pass the global context features through a fully-connected layer to output features with the same number of channels as the last layer of the segmentation network. We then add the features to the output of the segmentation network at each spatial location. (b) demonstrates how we encode the non-parametric prior information. For the spatial prior sampled with an $S \times S$ grid, we re-scale it to the same spatial dimension as the segmentation network output, then we propagate the priors through a convolutional layer. Finally, we add the outputs of the convolutional layer to the segmentation network output. For the global prior, we encode it using the same structure with feature encoding.}
   %\end{center}
   \vspace{-3mm}
   \label{figure: encoding}
\end{figure*}

%HERE
\vspace{1mm}
{\noindent {\bf Spatial Prior.}} 
Given a query image, we obtain the retrieval set $\{\mathbf{x}_1, \mathbf{x}_2, \ldots, \mathbf{x}_{K_p}\}$ with their annotated images $\{\mathbf{y}_1, \mathbf{y}_2, \ldots,  \mathbf{y}_{K_p}\}$. 
All the annotated images are first re-scaled to the same resolution and divided equally into $S \times S$ grids. Then we estimate the spatial prior as
\begin{equation}
\mathbf{P}_s[c,p,q] = \dfrac{1}{K_p}\sum_{k \in 1 \ldots K_p} N(\mathbf{y}_k[p,q],c),
\label{eqn: prior_s}
\end{equation}
where $N(\mathbf{y}_k[p,q],c)$ represents how many pixels are labeled as class $c$ within the specific block at the spatial coordinate $(p,q) \in S^2$ in the labeled image $\mathbf{y}_k$.
We can observe that $\mathbf{P}_s$ is an $C \times S \times S$ tensor, in which each location is a probability distribution with respect to all classes.
   
The spatial prior can be seen as a simplified version of local belief in the conventional non-parametric methods.
We estimate the probability in a lower resolution using spatial grids instead of superpixels, and we do not perform dense alignment such as SIFT-Flow~\cite{siftflow} or superpixel matching~\cite{superparsing} on the retrieval images to generate a detailed prediction.
This is because that our method already has the accurate local belief provided by the segmentation network, 
while the spatial prior information can provide a more consistent global configuration and eliminate false positives of local predictions. 
In addition, since we pass the prior information along with the deep features through convolution layers, the prior information can be propagated through the convolution operation, 
letting the network learn how to exploit additional cues through back propagation.
   
\vspace{1mm}
{\noindent {\bf Global Prior.}}
For {\em things} classes, we propose to utilize another prior information that is invariant with the spatial location and only estimates the existence of the object classes.
We denote such global prior as $\mathbf{P}_g$ which can be simply computed as
\begin{equation}
\mathbf{P}_g[c] = \sum_{k \in 1 \ldots K_p} \mathbf{N}(\mathbf{y}_k, c)/(h_k \times w_k \times K_p),
\label{eqn: prior_g}
\end{equation}
where $\mathbf{N}(\mathbf{y}_k, c)$ denotes the number of pixels in k-th retrieval image belonging to class $c$. 
We only compute the global prior on the {\em things} classes (e.g., person, animal, and chair) since the prior information of most {\em stuff} classes can be described accurately through the spatial prior in \eqref{eqn: prior_s}.

%========================================================================  
\section{Global Context Embedding with Networks}
With global context features generated by the global context network (Section \ref{section: GCN}) and the non-parametric prior information (Section \ref{section: prior}), 
we present how we apply both sets of context information for scene parsing.
   
Figure \ref{figure: encoding} shows the two modules that encode the global context features and non-parametric priors, respectively. 
To encode features, a naive approach is to duplicate the context features at each spatial location and concatenate them with feature maps of the last layer in the segmentation network. 
However, considering the convolutional layer with the kernel size of $1 \times 1$, it is mathematically equivalent to passing the context embedded features through a fully-connected layer, which has the same output channel number as the feature map of the segmentation network.
Then the output vector is added to each spatial location in the segmentation network as a bias term before the non-linearity function.
This modification can save memory for storing the duplicate feature vectors and accelerate the computing process. 
Furthermore, it makes the module easily applicable to any network without network surgery if it needs to be initialized from a pre-trained model.

For the prior encoding network, we encode the global prior using the same structure in the feature encoding network.
Since the spatial prior is not a $1$-dimensional vector, we first perform in-network resizing on the spatial prior with a bilinear kernel to match the feature map size. 
After resizing, we propagate the spatial prior through a convolutional layer and add the output to the feature map of the segmentation network.

%image row macros
\newcommand{\imgrow}[1]{            
   \includegraphics[width = 0.15\linewidth, height=0.15\linewidth]
   {{imgs/results/image/ADE_val_0000#1.jpg}} & \hspace{-0.22cm}
   \includegraphics[width = 0.15\linewidth, height=0.15\linewidth]
   {{imgs/results/annotation/ADE_val_0000#1.png}} & \hspace{-0.22cm}
   \vline \vline & \hspace{-0.22cm}
   \includegraphics[width = 0.15\linewidth, height=0.15\linewidth]
   {{imgs/results/org_80000/ADE_val_0000#1.png}} & \hspace{-0.22cm}
   \includegraphics[width = 0.15\linewidth, height=0.15\linewidth]
   {{imgs/results/f_iccv_513_ar_flip_nosm_70000_color/ADE_val_0000#1.png}} & \hspace{-0.22cm}
   \includegraphics[width = 0.15\linewidth, height=0.15\linewidth]
   {{imgs/results/s+g_iccv_513_ar_flip_nosm_70000_color/ADE_val_0000#1.png}} & \hspace{-0.22cm}
   \includegraphics[width = 0.15\linewidth, height=0.15\linewidth]
   {{imgs/results/s+g+f_iccv_513_ar_flip_nosm_96000_color/ADE_val_0000#1.png}} &  \\
}

\begin{figure*}[!t]
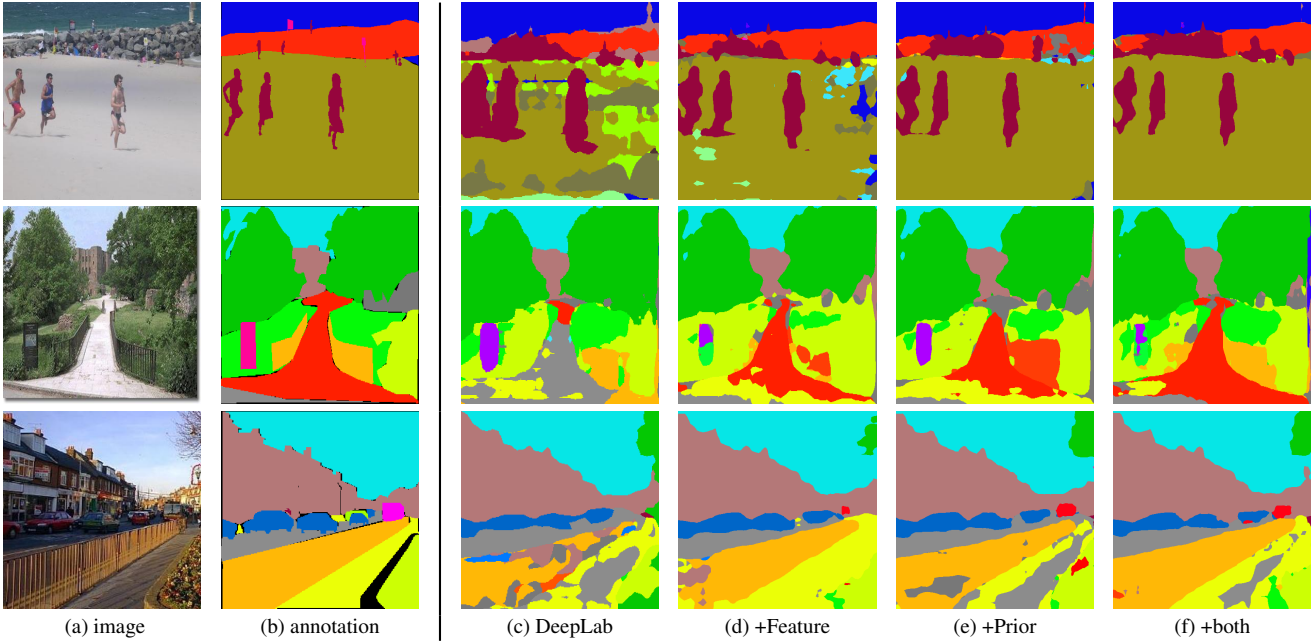
\footnotesize
   \footnotesize
   \begin{center}
      \begin{tabular}{@{}ccccccccc@{}}
         
         \imgrow{0744}
         %\imgrow{0099}
         %\imgrow{0117}
         %\imgrow{0125}
         %\imgrow{0661}
         \imgrow{0682}
         %\imgrow{0921}
         %\imgrow{0757}
         %\imgrow{0777}
         \imgrow{0867}
         %\imgrow{1078}
         
         \vspace{0.1cm} 
         
         (a) image & \hspace{-0.22cm} 
         (b) annotation & \hspace{-0.22cm} 
         \vline \vline & \hspace{-0.22cm} 
         (c) DeepLab & \hspace{-0.22cm}
         (d) +Feature & \hspace{-0.22cm}
         (e) +Prior & \hspace{-0.22cm}
         (f) +both & \hspace{-0.22cm} \\
         
      \end{tabular}
   \end{center}
   \vspace{-0.3cm}
   \caption{Representative scene parsing results from the ADE20k dataset.}
   \vspace{-3mm}
   \label{figure: mit}
\end{figure*}

\section{Experimental Results}

{\noindent {\bf Implementation Details.}} 
We use the caffe toolbox~\cite{jia2014caffe} to train our models with TitanX GPUs
with the mini-batch stochastic gradient descent. 
To learn the global context network, we train the network in Figure \ref{figure: siamese} with mini-batch size $16$.
Each mini-batch contains 8 positive and 8 negative pairs. We set the nearest neighbor number $K_a$ in equation (\ref{eqn: binarize}) as $10$ for generating the positive pairs.
The initial learning rate is set to $0.001$ which is reduced to $0.0001$ after $100,000$ iterations. 
The momentum is set as $0.9$ with weight decay $0.0005$. We initialize the model with the pre-trained VGG-16 model~\cite{vgg} on ImageNet and train it for $150,000$ iterations.
   
For the full system training as depicted in Figure \ref{figure: system}, we perform experiments using two baseline models: FCN-8s~\cite{fcn} and DeepLab-ResNet101~\cite{deeplab}.
For FCN-8s, we train the network with unnormalized softmax with fixed learning rate $1\mathrm{e}{-10}$.
The model is initialized with the pre-trained VGG-16 model on ImageNet. We apply the heavy-learning scheme in ~\cite{fcn} where we set batch size to $1$ and momentum to $0.99$.
For DeepLab-ResNet101, we follow the learning rate policy in their paper where initial learning rate is set to $2.5\mathrm{e}{-4}$ and is lowered by the polynomial scheme with power $0.9$ and max iteration $160000$.
   
We set input crop size as $384$ for FCN and $385$ for DeepLab since it requires the input edges to be in the form of $32N+1$. 
For data augmentation, we use random-mirroring for both baseline models.
When training the DeepLab network, we additionally apply random scaling with a choice of 5 scales \{0.5, 0.75, 1.0, 1.25, 1.5\} for each iteration, and crop the image to the input size with a random position. 
Note that we also perform the same scaling and cropping on the spatial prior. For DeepLab, since it consists of 3 branches with different resolution, we apply the encoding module separately on all the branches and resize the spatial prior accordingly.
For evaluation, the test image is resized to have the longer edge as $513$ ($384$ for FCN), and we do not perform multi-scale testing.
We also apply dense CRF as the post-processing module but find there is no significant improvement ($\pm 0.5\%$) since the CRF can only improve the boundary alignment. Also, it will take another 10-40 seconds to optimize the dense CRF for one image. 
The code and model are available at \url{https://github.com/hfslyc/GCPNet}
%\vspace{-0.2cm}
\setlength{\tabcolsep}{12pt}
\begin{table}[t]
   \caption{ $K_a$ in constructing affinity matrix with $K_p=5$}
   
   \vspace{0mm}
   \label{table: siamese_k}
   \centering
   \begin{tabular}{ccccc}
      
      \hline
      $K_a$ & 5 & \textbf{10} & 15 & 20 \\
      \hline
      %recall & 73.89& \textbf{74.76} & 74.24 & 74.49 \\
      %precision & 27.52 & \textbf{27.96} & 28.10 & 28.02\\
      $F_2$ & 0.5527 & \textbf{0.5601} & 0.5589 & 0.5594 \\
      \hline
   \end{tabular}
   \vspace{-1mm}     
\end{table}

\setlength{\tabcolsep}{12pt}
\begin{table}[t]
   \caption{$K_p$ in prior generation with $K_a=10$}
   
   \vspace{0mm}
   \label{table: prior_k}
   \centering
   \begin{tabular}{ccccc}
      
      \hline
      $K_p$ & 1 & 3 & \textbf{5} & 10 \\
      \hline
      %recall & 43.97 & 64.54 & \textbf{74.76} & 82.82 \\
      %precision & 45.16 & 32.48 & \textbf{27.96} & 21.12\\
      $F_2$ & 0.4420 & 0.5390 & \textbf{0.5601} & 0.5228 \\
      \hline
   \end{tabular}
   \vspace{0mm}
\end{table}

\vspace{3mm}
\noindent {\bf Hyperparameters.}
We analyze two important hyper parameters in the proposed method: 
1) $K_a$ for constructing affinity matrix in the global context network training in \eqref{eqn: binarize};
2) $K_p$ as the amount of retrieved images for generating the priors in \eqref{eqn: prior_s} and \eqref{eqn: prior_g}. 
We choose both values based on the quality of the scene retrieval results in Table~\ref{table: siamese_k} and Table~\ref{table: prior_k}. 

The retrieved images should contain most classes that appear in the query image (high recall) and few irrelevant classes. 
Thus, we treat the retrieval results as a multi-label classification problem and evaluate the $F_2$ score with different parameters on the ADE20K validation set, where $F_\beta = (1 + \beta^2) \cdot precision \cdot recall/ (\beta^2 \cdot precision + recall)$. We choose $F_2$ since we prefer results with the higher recall.

Table \ref{table: siamese_k} shows the results with different values of $K_a$.
The global context network performs well for a wide range of values of $K_a$, and we choose $K_a=10$ with the highest $F_2$ score.
Table~\ref{table: prior_k} shows the sensitivity analysis on $K_p$.
The proposed method performs best when $K_p$ is set to 5.

\vspace{5mm}
\begin{figure}[t]
   \scriptsize
   \centering
   \begin{tabular}{cc}
      \hspace{-1.0cm}
      \begin{adjustbox}{valign=c}
         \begin{tabular}{c}
            \includegraphics[width=0.25\linewidth, height=0.25\linewidth]{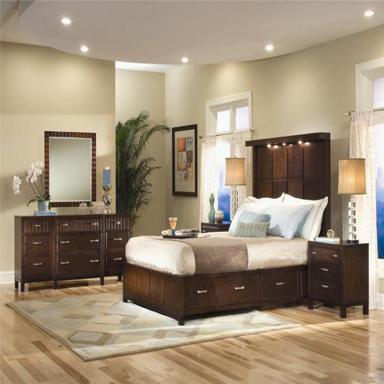}
            \\
            query image
         \end{tabular}
      \end{adjustbox}
      \hspace{-0.3cm}
      \begin{adjustbox}{valign=c}
         \begin{tabular}{cccc}
            \includegraphics[width=0.15\linewidth, height=0.15\linewidth]{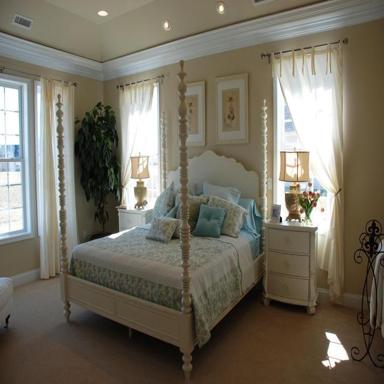} \hspace{-0.7cm} &
            \includegraphics[width=0.15\linewidth, height=0.15\linewidth]{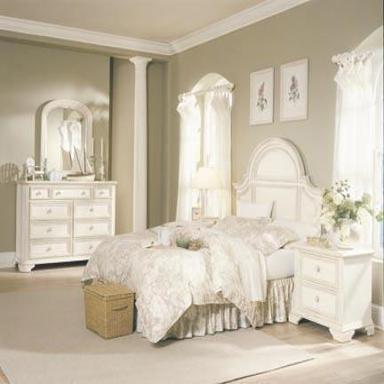} \hspace{-0.7cm} &
            \includegraphics[width=0.15\linewidth, height=0.15\linewidth]{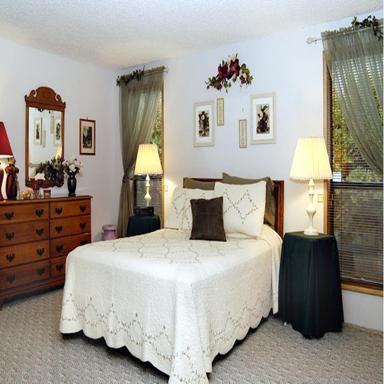} \hspace{-0.7cm} &
            \includegraphics[width=0.15\linewidth, height=0.15\linewidth]{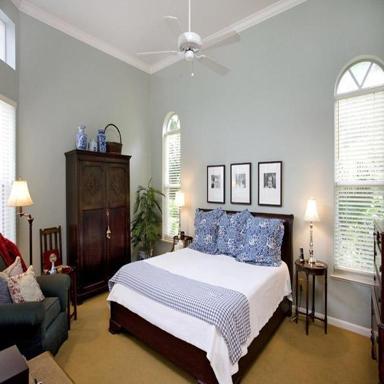}
            
            \\
            \includegraphics[width=0.15\linewidth, height=0.15\linewidth]{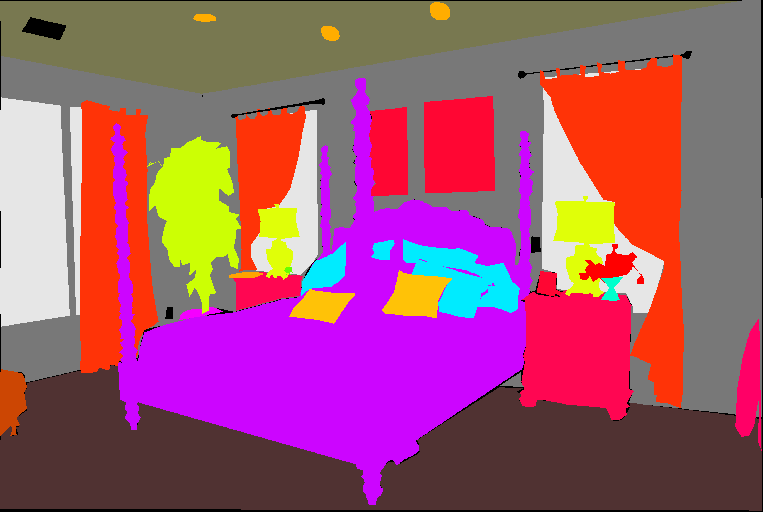} \hspace{-0.7cm} &
            \includegraphics[width=0.15\linewidth, height=0.15\linewidth]{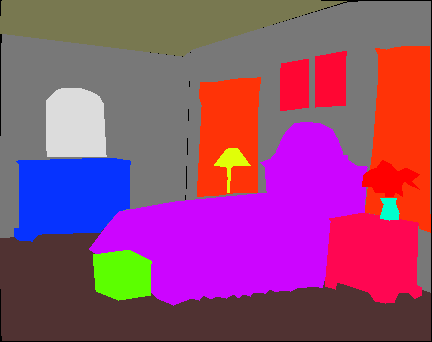} \hspace{-0.7cm} &
            \includegraphics[width=0.15\linewidth, height=0.15\linewidth]{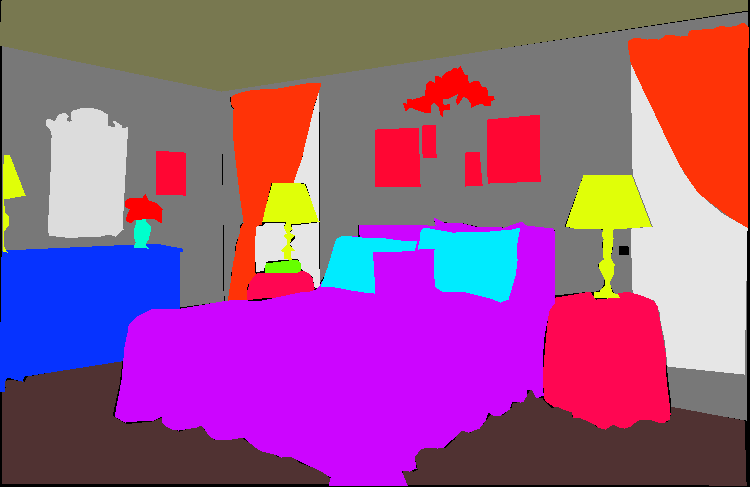} \hspace{-0.7cm} &
            \includegraphics[width=0.15\linewidth, height=0.15\linewidth]{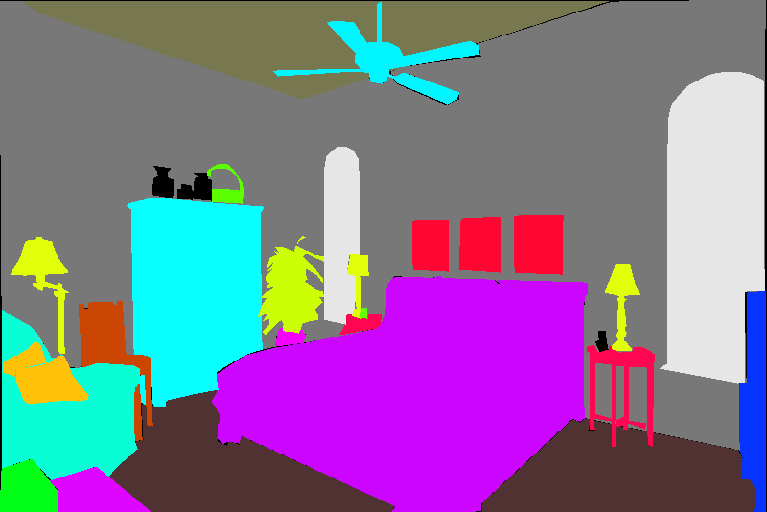} 
            \\ 
            \multicolumn{4}{c}{retrieval set using global context features}
         \end{tabular}
      \end{adjustbox}
   \end{tabular}

   \begin{tabular}{cc}
      \hspace{-1.0cm}
      \begin{adjustbox}{valign=c}
         \begin{tabular}{c}
            \includegraphics[width=0.25\linewidth, height=0.25\linewidth]{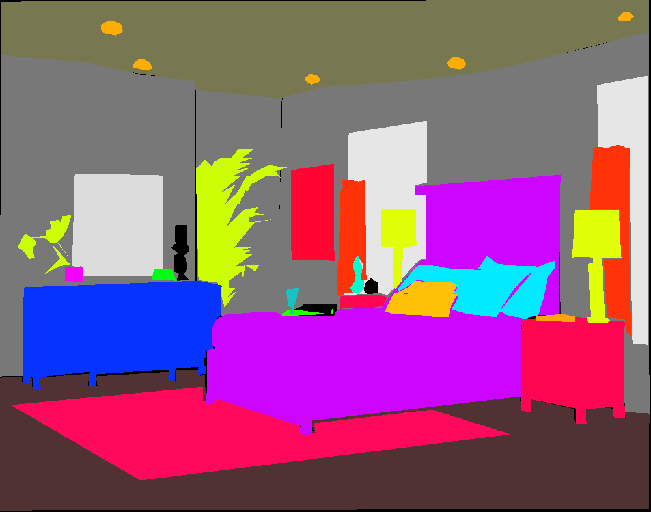}
            \\
            annotation
         \end{tabular}
      \end{adjustbox}
      \hspace{-0.3cm}
      \begin{adjustbox}{valign=c}
         \begin{tabular}{cccc}
            \includegraphics[width=0.15\linewidth, height=0.15\linewidth]{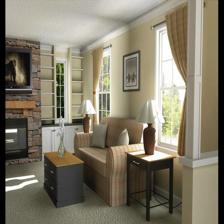} \hspace{-0.7cm} &
            \includegraphics[width=0.15\linewidth, height=0.15\linewidth]{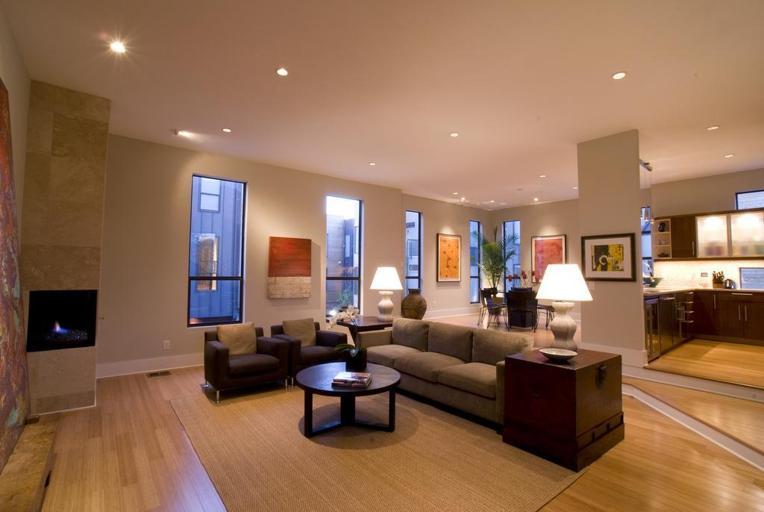} \hspace{-0.7cm} &
            \includegraphics[width=0.15\linewidth, height=0.15\linewidth]{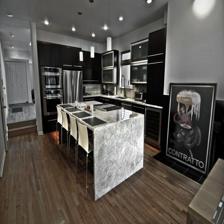} \hspace{-0.7cm} &
            \includegraphics[width=0.15\linewidth, height=0.15\linewidth]{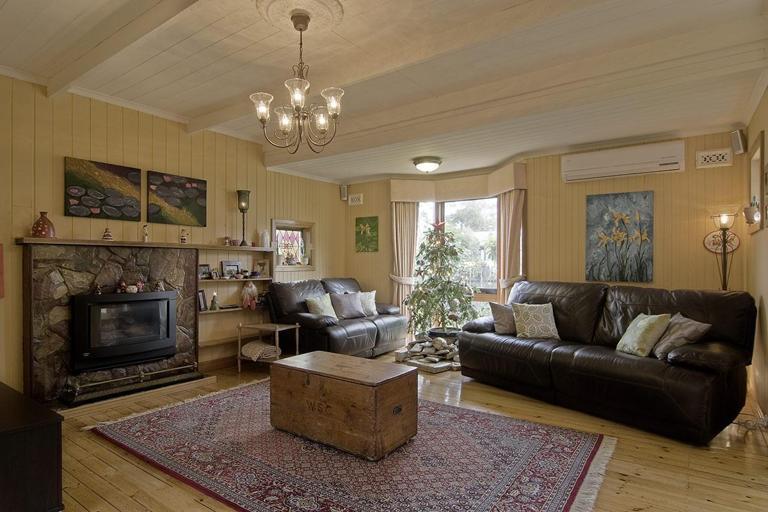}
            
            \\
            \includegraphics[width=0.15\linewidth, height=0.15\linewidth]{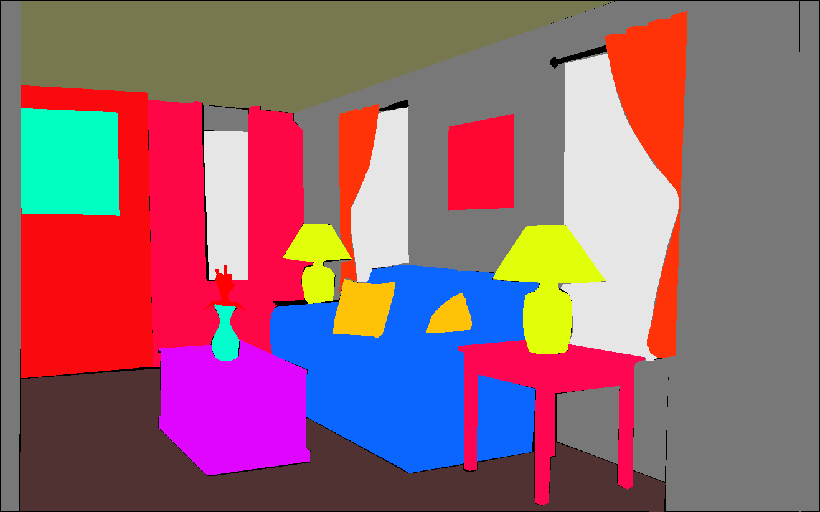} \hspace{-0.7cm} &
            \includegraphics[width=0.15\linewidth, height=0.15\linewidth]{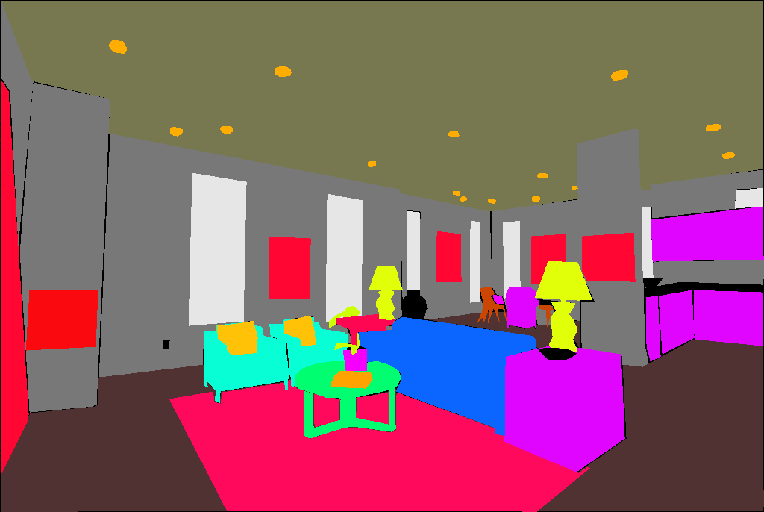} \hspace{-0.7cm} &
            \includegraphics[width=0.15\linewidth, height=0.15\linewidth]{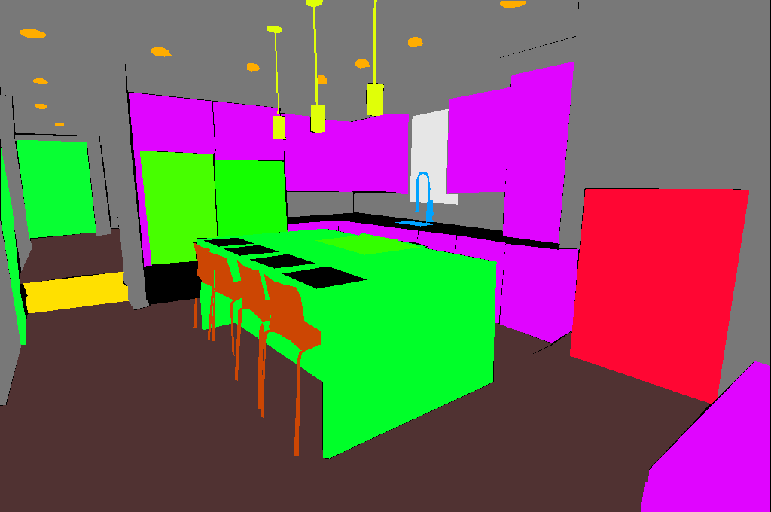} \hspace{-0.7cm} &
            \includegraphics[width=0.15\linewidth, height=0.15\linewidth]{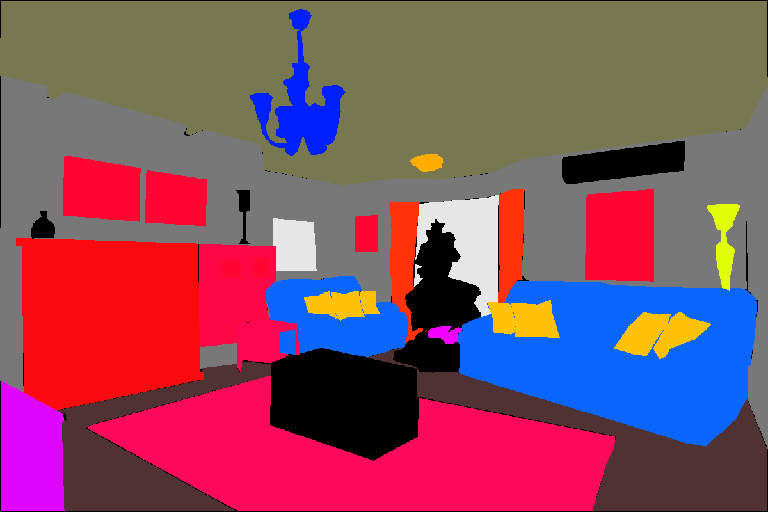} 
            \\ 
            \multicolumn{4}{c}{retrieval set using VGG-16 features pretrained on ImageNet}
         \end{tabular}
      \end{adjustbox}
   \end{tabular}
   %\vspace{0.1cm}
   \caption{Comparison of retrieval results between VGG-16 feature and our trained global context feature. We note that none of the results using the VGG-16 features contains the ``bed'' label.}
   \vspace{-1mm}  
   \label{figure: siamese_retrieval_vgg}
\end{figure}

\vspace{-2mm}  
\noindent {\bf Quantitative Evaluation of Global Context Features.}

To validate our learned global context features, we compare the results using our features to the ones that directly use \textbf{fc7} of the VGG-16 network pre-trained on ImageNet with the $F_2$ score in Table \ref{table: siamese_k}. 
The one using VGG-16 features only achieves 0.2524 with $K_p=5$, which is substantially lower than our global context features (0.5601). 
It shows that the global context features can learn useful scene semantic information from the proposed siamese training. 
Figure \ref{figure: siamese_retrieval_vgg} shows one example of the retrieval results with these two features.

\vspace{1mm}
\noindent {\bf Complexity Analysis}
For evaluating a single test image, our method takes 1.4 seconds: 0.4s (NN search) + 0.3s 
(prior generation) + 0.7s (Network forward) on a single GPU. 
The prior encoding module (NN search + prior generation) introduces 0.7s 
overhead on CPU, which can be minimized via GPU acceleration. 
Our module is efficient compared to other non-parametric methods such as ~\cite{siftflow, superparsing}, since we perform retrieval at the image level with pre-computed context features while others perform it either at the pixel or superpixel level.
%
%Compared to the size of the caffe model weight (650 MB), we only need extra 300 MB to store the precomputed context features and the coarse annotations of the ADE 20K training set, which can be easily compressed to be more compact.

\vspace{1mm}   
{\noindent {\bf MIT ADE20K Dataset.}} 
We first validate our methods on the recently published MIT ADE20K dataset~\cite{zhou2016semantic}. 
The dataset consists of 20,000 image in the \emph{train} set and 2,000 images in the \emph{validation} set. 
There are total 150 semantic classes, in which 35 classes belong to {\em stuff} classes, and 115 classes belong to {\em things} classes. 
The dataset is considered as one of the most challenging scene parsing datasets due to its scene variety and numerous annotated object instances.
We first train the global context network on the \emph{train} set with $K_a$ as 10 to retrieve positive pairs. 
The negative pairs are sampled with the strategy mentioned in Section \ref{section: GCN} where $N$ is set to $10000$. 
The spatial and global prior is generated with the nearest neighbor parameter $K_p$ as 5 and spatial grid $50 \times 50$. 
We compare our method with several state-of-the-art models: FCN~\cite{fcn}, DilatedNet~\cite{dilated}, DeepLab~\cite{deeplab} and the cascade model presented along with the dataset~\cite{zhou2016semantic}.
We show the performance comparison in Table \ref{table: mit}. 
By applying the feature encoding and the prior encoding on the FCN-8s model, we get $3.09\%$ and $3.53\%$ mean-IU improvement, respectively. 
We also apply our modules on the DeepLab model~\cite{deeplab} based on ResNet-101~\cite{resnet}. 
By applying the feature encoding, we have $3.12\%$ mean-IU improvement over the baseline.
Prior encoding brings similar improvement with $3.42\%$ difference. We also combine both modules and perform joint training and achieve $38.37\%$ mean-IU with $4.43\%$ improvement over the baseline model. 
We also show the qualitative comparison of our modules in Figure~\ref{figure: mit}.
In addition, if we define the most frequent 30 classes as the common classes and the rest of them as the rare classes, the average improvement of mean IU is $3.13\%$ for common classes and $4.64\%$ for rare classes. 
This shows that our method can improve both common and rare classes without sacrificing the other.

\setlength{\tabcolsep}{6pt}
\begin{table}[t]
   \caption{Results on the MIT ADE20k \emph{validation} set.}
   \vspace{2mm}
   \label{table: mit}
   \centering
   \begin{tabular}{lcc}
      \hline
      Methods & Pixel Accuracy & Mean IU \\
      \hline
      FCN-8s~\cite{zhou2016semantic} & 71.32 & 29.39 \\
      
      DilatedNet~\cite{zhou2016semantic} & 73.55 & 32.31 \\
      
      Cascade-DilatedNet~\cite{zhou2016semantic} & 74.52 & 34.9 \\
      
      \hline
      FCN-8s + Feature & 74.47 & 32.48 \\
      
      FCN-8s + Prior & 75.00 & 32.92 \\
      \hline
      DeepLab~\cite{deeplab} & 75.80 & 33.94 \\
      
      DeepLab + Feature & 77.47 & 37.06 \\         
      
      %DeepLab + Feature + CRF & 77.30 & 36.28 \\
      
      DeepLab + Prior & \bf77.94 & 37.46 \\
      
      %DeepLab + Prior + CRF & \bf77.92 & 36.49 \\
      \hline
      DeepLab + Feature + Prior & {77.76} & \bf{38.37} \\
      \hline
   \end{tabular}
   \vspace{2mm}
\end{table}

\vspace{2mm}
{\noindent {\bf PASCAL Context Dataset.}
We also evaluate our method on the PASCAL Context dataset~\cite{pascal_context}.
It consists of 60 classes with 4998 images in the \emph{train} set and 5105 images in the \emph{val} set. 
The performance comparison is shown in Table \ref{table: pascal_context}.
We apply both global context feature encoding and prior encoding on top of the baseline model. 
Different from the models on the MIT ADE20k dataset, we generate the spatial and global prior information on both {\em stuff} and {\em things} classes.
Both the baseline and our models are trained 20000 iterations with batch size 10.
Our method achieves $73.80\%$ pixel accuracy and $46.52\%$ mean IU, which has the favorable performance against state-of-the-art methods.

\setlength{\tabcolsep}{6pt}
\begin{table}[t]
   \caption{Results on the PASCAL-Context \emph{validation} set.}
   \vspace{2mm}
   \label{table: pascal_context}
   \centering
   \begin{tabular}{lcc}
      \hline
      Methods & Pixel Accuracy & Mean IU \\
      \hline
      $\text{O}_2\text{P}$~\cite{carreira2012semantic} & N/A & 18.1\\
      CFM~\cite{dai2015convolutional} & N/A & 34.4\\
      FCN-8s~\cite{fcn} & N/A & 37.78\\
      CRF-RNN~\cite{crfasrnn} & N/A & 39.28\\
      BoxSup~\cite{dai2015boxsup} & N/A & 40.5\\
      HO\_CRF~\cite{arnab2016higher} & N/A & 41.3\\
      \hline
      DeepLab & 71.57 & 44.38\\
      %DeepLab + CRF & 70.20 & 43.12\\
      DeepLab + Feature + Prior & \bf73.80 & \bf46.52\\
      %DeepLab + All & \bf{71.75} & \bf{44.03} \\
      \hline
      
   \end{tabular}
   \vspace{-3mm}
\end{table}

\vspace{-2mm}
\section{Conclusion}
\vspace{-2mm}
We present a novel scene parsing system that exploits the global context embedding representation. 
Through learning from the scene similarity, we generate a global context representation for each image to aid the segmentation network. 
We show that the proposed algorithm, which consists of feature encoding and non-parametric prior encoding, can be applied to most state-of-the-art segmentation networks. 
Based on the proposed method, we achieve significant improvement on both the challenging MIT ADE20K dataset and the PASCAL Context dataset.

\vspace{-2mm}
\section*{Acknowledgments}
\vspace{-2mm}
This work is supported in part by the NSF CAREER Grant \#1149783, gifts from Adobe and NVIDIA.

\clearpage
{\small
   \bibliographystyle{ieee}
   \bibliography{scene_parsing}
}

\end{document}